%% file: main.tex
\documentclass[sigconf, nonacm]{acmart}

\usepackage{pvldb}

\usepackage{xurl}
\renewcommand\vldbdoi{XX.XX/XXX.XX}
\renewcommand\vldbpages{XXX-XXX}
\renewcommand\vldbavailabilityurl{%
https://drive.google.com/drive/folders/1szHWGzLpmgCG_TCnU4W4I7Jom9yzxA63%
}

\usepackage{amsmath}
\usepackage{bm}
\usepackage{graphicx}
\usepackage{etoolbox}
\usepackage{enumitem}
\usepackage{algorithmic}
\usepackage{multirow}
\usepackage{amsthm}
\usepackage{booktabs}
\usepackage{xspace}
\usepackage{adjustbox}
\usepackage{amsmath,stmaryrd,pict2e,picture}
\usepackage{xcolor}
\usepackage{tikz}
\usepackage{listings}
\usepackage{bbold}
\usepackage{diagbox}
\usepackage{balance}
\usepackage{float}
\usepackage[noend,lined,boxed,vlined,ruled,linesnumbered]
{algorithm2e}
\usepackage{makecell}
\usepackage{array}
\usepackage{arydshln}
\usepackage{caption, subcaption}
\usepackage[labelformat=simple]{subcaption}

\newcommand{\subrow}[1]{{#1}}
\definecolor{sigteal}{HTML}{008C8C}
\newcommand{\sigstar}[1]{\textcolor{sigteal}{#1}}

\newcommand{\ci}[1]{\textcolor{gray}{\small#1}}
\newcommand{\cisub}[1]{\textcolor{gray}{\fontsize{6}{7}\selectfont #1}}

\definecolor{ToolBlue}{RGB}{44, 95, 140}

\definecolor{TrainOrange}{RGB}{44, 95, 140}

\definecolor{AllBlue}{RGB}{52, 73, 94}

\definecolor{AllHighlight}{RGB}{235, 242, 250}

\newcommand{\toolimp}[1]{\textcolor{ToolBlue}{\footnotesize$\bm{\uparrow}_{\bm{\textsc{Tool}}}\,\bm{#1}$}}
\newcommand{\trainimp}[1]{\textcolor{TrainOrange}{\footnotesize$\bm{\uparrow}_{\bm{\textsc{Train}}}\,\bm{#1}$}}
\newcommand{\allimp}[1]{%
  \colorbox{AllHighlight}{\textcolor{black}{($\bm{\uparrow}\bm{#1}$})}%
}

\newcolumntype{C}[1]{>{\centering\arraybackslash}p{#1}}

\definecolor{myyellow}{HTML}{fcefd2}
\definecolor{mydarkyellow}{HTML}{FAD02E}

\definecolor{mygreen}{HTML}{d2fce6}
\definecolor{mypurple}{HTML}{f2d2fc}
\definecolor{peach}{HTML}{FFDAB9}
\definecolor{fpred}{RGB}{255,99,99}

\SetKwInOut{Input}{input}
\SetKwInOut{Output}{output}

\newcommand{\codeq}[1]{{\tt {\small ``#1''}}}

\newcommand{\code}[1]{\texttt{\small #1}}

\newcommand{\ignore}[1]{}

\newtheorem{df}{Definition}
\newtheorem{ex}{Example}
\newtheorem{pr}{Proposition}
\newtheorem{lemm}{Lemma}
\newtheorem{lem}{Theorem}

\newenvironment{example}{\begin{ex} \nopagebreak
\begin{rm}}{{\hfill$\Box$}\end{rm}\end{ex}}

\newenvironment{definition}{\begin{df} \nopagebreak
\begin{rm}}{{}\end{rm}\end{df}}

\newcommand{\total}{24\xspace}

\newtoggle{full}
\toggletrue{full}

\definecolor{tool_arrow}{HTML}{649393}
\definecolor{posttrain_arrow}{HTML}{7A4E5D}
\definecolor{visualization_pink}{HTML}{EA6B66}
\definecolor{export_box}{HTML}{67AB9F}
\definecolor{fact_table_box}{HTML}{D79B00}

\newcommand\problem{\textsc{End-to-end BI}\xspace}
\newcommand\bench{\textsc{BI-Bench}\xspace}
\newcommand\sys{\textsc{BI-Agent}\xspace}
\newcommand\tools{\textsc{Tools}\xspace}

\newcommand{\minihead}[1]{{\vspace{.5em}\noindent\textbf{#1} }}

\begin{document}

\title{\sys and \bench: \\Towards Automating End-to-End Business Intelligence}

\author{Chuxuan Hu}
\authornote{Part of work done while at Microsoft.}
\affiliation{%
  \institution{UIUC}
  \city{Urbana}
  \state{Illinois}
  \country{USA}
}

\author{Yeye He}
\authornote{Correspondence: chuxuan3@illinois.edu, yeyehe@microsoft.com}
\affiliation{%
  \institution{Microsoft Research}
  \city{Redmond}
  \state{Washington}
  \country{USA}
}

\author{Penny Zhou}
\affiliation{%
  \institution{Microsoft}
  \city{Redmond}
  \state{Washington}
  \country{USA}
}

\author{Wee Hyong Tok}
\affiliation{%
  \institution{Microsoft}
  \city{Redmond}
  \state{Washington}
  \country{USA}
}

\author{Daniel Kang}
\affiliation{%
  \institution{UIUC}
  \city{Urbana}
  \state{Illinois}
  \country{USA}
}

\author{Surajit Chaudhuri}
\affiliation{%
  \institution{Microsoft Research}
  \city{Redmond}
  \state{Washington}
  \country{USA}
}

\input{tex/0-Abstract}

\maketitle
\pagestyle{plain}


\begin{sloppy}

\input{tex/1-Introduction}

    \input{tex/2-Related}

\input{tex/3-Problem}
\input{tex/4-Data}
\input{tex/5-Method}
\input{tex/6-experiment}
\input{tex/7-Conclusions}

\clearpage


\bibliographystyle{ACM-Reference-Format}
\bibliography{BI-agent}
\clearpage

\iftoggle{full}
{
    \appendix

}

\end{sloppy}

\end{document}

%% file: tex/0-Abstract.tex
\begin{abstract}
Business intelligence (BI) is a cornerstone of enterprise decision-making and is widely used by enterprise users in software such as Power BI and Tableau. In traditional BI workflows, users need to prepare data by (1) identifying relevant tables, (2) performing data transformations, and (3) building join relationships, before they can (4) answer their business questions. These steps can be complex and time-consuming, making BI challenging.

Given the strong capabilities of large language models (LLMs) in working with data, we study their ability to answer BI questions end-to-end, without requiring users to manually perform the tedious preparation steps.
To do this, we harvest a large collection of real-world BI projects from public sources, and manually extract pairs of (\emph{business questions}, \emph{ground-truth answers}) from real user dashboards. The resulting benchmark, \bench, is the first benchmark to systematically study LLMs' ability on end-to-end BI.

We find that even frontier LLMs perform poorly on \bench, with less than 50\% accuracy.
To address their limitations, we design a tool-augmented \sys, that decomposes BI workflows into subtasks on structured data, such as search, join, and transform, and orchestrates specialized data management methods across BI stages. Furthermore, we develop a post-training framework that synthesizes training trajectories from real BI projects, enabling \sys to be further post-trained using both supervised fine-tuning (SFT) and reinforcement learning (RL).
\sys achieves substantial accuracy gains of up to 40 percentage points on \bench with vanilla LLMs, and post-trained \sys yields gains of up to 30 points, both of which are highly statistically significant.
Our results highlight the importance of combining tool-augmented reasoning with domain-specific post-training in complex BI workflows, and point to promising directions for future research.
We release our code and data at \url{https://github.com/Hu-Chuxuan/bi-agent}.


\end{abstract}

%% file: tex/1-Introduction.tex


\section{Introduction}
\label{sec:intro}

\begin{figure}[t]
    \graphicspath{{figures/}}
    \centering
    \includegraphics[width=\columnwidth]{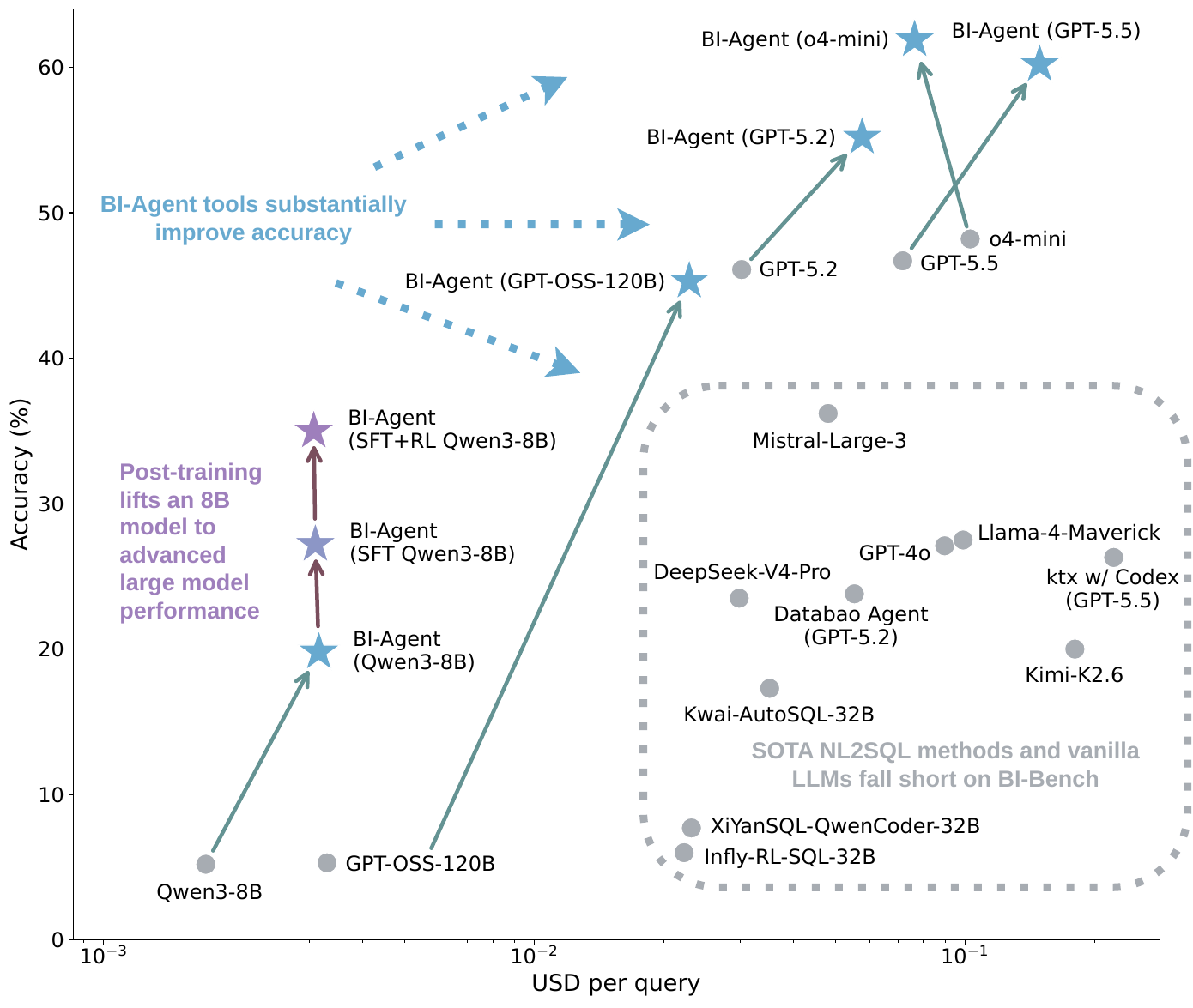}
    \caption{Quality and cost comparison of \sys vs. other methods, including frontier models (e.g., GPT-5.5) and SOTA NL2SQL systems (e.g., Kwai-AutoSQL and ktx), on \bench using SQL. Complete results on \total models and systems can be found in Tables~\ref{tab:main_res_tool} and ~\ref{tab:main_res_post_train}. \sys using data-management tools significantly improves frontier models by over 10 percentage points on average (shown by \textbf{\textcolor{tool_arrow}{teal}} arrows); while \sys with post-training (shown by \textbf{\textcolor{posttrain_arrow}{purple}} arrows) enables Qwen3-8B to achieve  quality comparable to much larger frontier models, at substantially lower costs (up to 50$\times$ cheaper). 
    }
    \label{fig:cost-acc-comp}%
\end{figure}

\begin{figure*}[t!]
    \graphicspath{{figures/}}
    \centering
    \includegraphics[width=\textwidth]{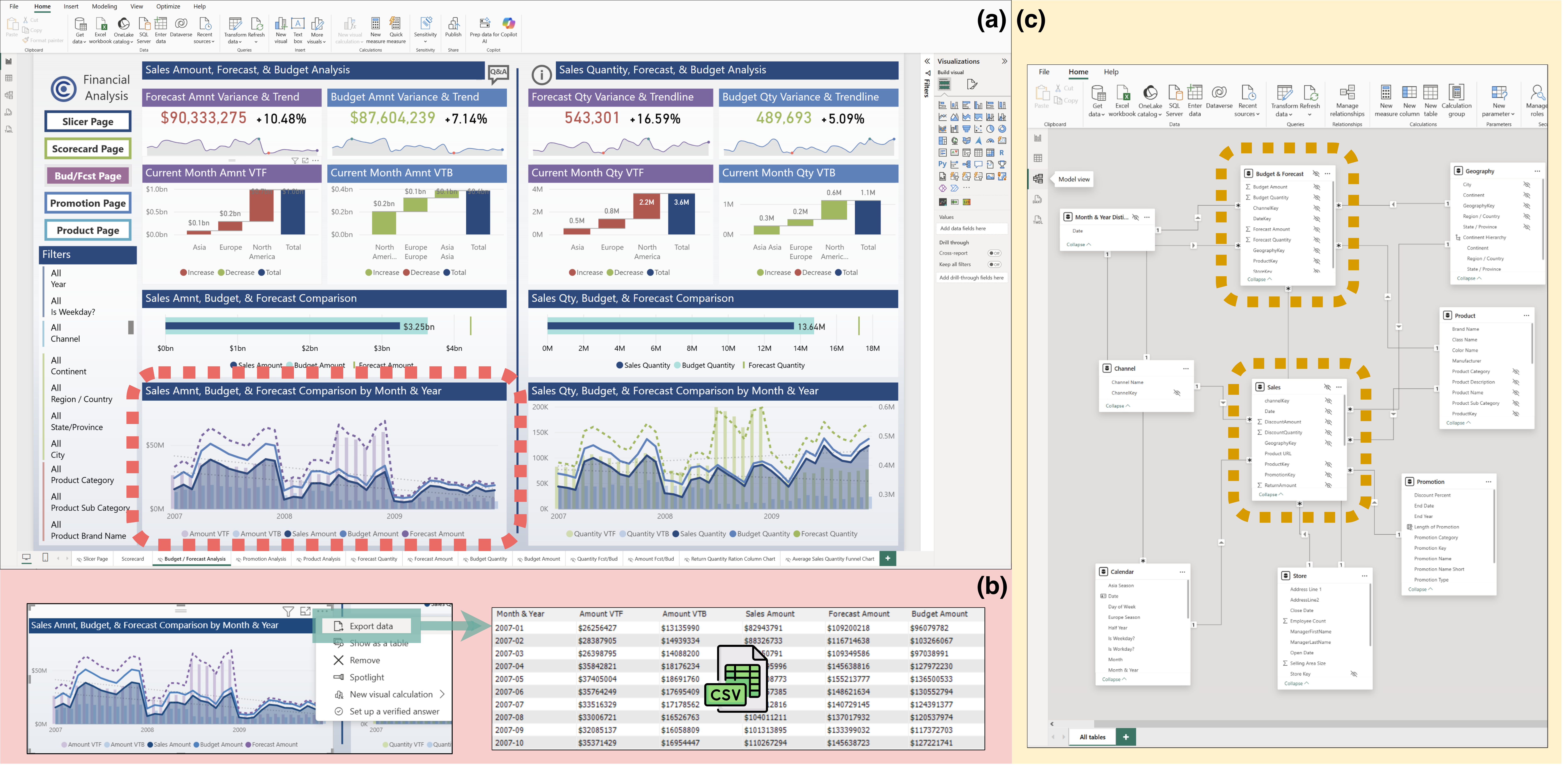}
    \caption{(\underline{a}): An example BI dashboard built by users in the wild, extracted from a real Power BI project file (.pbix).
    For \bench, we manually select a visualization whose title can accurately describe user intent: e.g., the visualization in the dashed \textcolor{visualization_pink}{pink} box corresponds to a user question $Q$ = \codeq{Sales amount, budget, and forecast comparison by month and year}.
    (\underline{b}): We directly retrieve the answer $A(Q)$ for $Q$ by exporting the underlying result table of the selected visualization as a CSV file through the ``Export data'' menu option (in the \textcolor{export_box}{green} box). This $(Q, A(Q))$ pair becomes the basis of a test case in \bench.
    (\underline{c}): The input data tables of this BI project in a visual data model view. Note that this schema graph follows a ``multi-snowflake'' pattern common in BI, with the ``Budget \& Forecast'' and ``Sales'' tables at the center serving as ``fact tables'' (in \textcolor{fact_table_box}{orange} boxes), and the remaining 7 tables serving as ``dimension tables''. 
    }
    \label{fig:screen-shots}%
\end{figure*}

Business Intelligence (BI) plays a crucial role in empowering modern enterprises to make informed data-driven decisions, and has grown into a multi-billion-dollar business~\cite{gartner-bi, forrester-bi}. Popular BI software, such as Power BI~\cite{pbi} and Tableau~\cite{tableau}, is used by over 100K organizations worldwide for decision-making~\cite{bi-usage-1, bi-usage-2}, underscoring the importance of BI in today’s data-driven enterprises.

\minihead{Traditional workflows of BI.}
Existing BI platforms, such as Power BI and Tableau, require users to go through fixed, multi-stage workflows to prepare raw data before carrying out the actual BI analysis, as documented in their official tutorials~\cite{tableau-workflows, pbi-workflows}. 

Figure~\ref{fig:screen-shots}(a) shows a real-world BI dashboard created by users in the wild using Power BI, designed to answer a wide range of business questions.
To build this and answer business questions, users have to go through a multi-step workflow: (\textbf{S1}) browse and identify relevant tables within the project (Figure~\ref{fig:screen-shots}(c) shows a small subset of the many tables available in the project that users have to pick and choose from); (\textbf{S2}) perform data transformations to make resulting tables suitable for analysis, often using vendor proprietary DSLs~\cite{m-dsl} or GUI~\cite{tableau-prep};  (\textbf{S3})  define join relationships between these tables, shown as join edges in Figure~\ref{fig:screen-shots}(c), to enable cross-table analysis. Only after these steps can (\textbf{S4}) the actual BI analysis be performed using dashboards in these existing BI platforms. 

\iftoggle{full}
{
    Tutorials from official sources, such as~\cite{tableau-workflows, pbi-workflows}, provide detailed, step-by-step instructions for completing these stages within the BI workflows for Power BI and Tableau,  mirroring the process above.
}
{
}

While these steps are powerful and flexible, they are complex for non-technical enterprise users \cite{8440815}. For instance, Figure~\ref{fig:screen-shots}(c) shows a subset of the tables in a BI project;
completing \textbf{S3} requires users to inspect these tables and define their join relationships. Similarly, completing \textbf{S2} requires users
to program transformation logic in vendor-provided DSLs. 
Both tasks are challenging without 
database 
or programming expertise.

Despite the user-friendly interfaces of modern BI tools, executing the full BI workflow (to select, transform, join, and analyze data) remains a key pain point for non-technical enterprise users.


\minihead{\bench: Building a benchmark for real-world end-to-end BI.}
Since large language models (LLMs) have shown strong capabilities in processing and analyzing data~\cite{gpt4, llama3, table-gpt, tablellama}, we start by studying their ability to automate end-to-end BI.

While there are many data analysis benchmarks, especially in the NL2SQL space (e.g.,~\cite{bird, wikisql, spider2}), none specifically target the end-to-end challenges in real BI workflows. To address this gap, we crawl a large collection of 
real-world BI projects (.pbix files) from publicly accessible webpages identified through a web search engine index, and manually extract/curate real business questions that are asked in these BI projects,
to construct the first benchmark for end-to-end BI that we call \bench.

Figure~\ref{fig:screen-shots} shows one example from this collection of real BI projects. Figure~\ref{fig:screen-shots}(a) is a final dashboard created by users, and Figure~\ref{fig:screen-shots}(c) displays a subset of its underlying data tables, including their schemas and join relationships.

To build \bench, we first sample a real BI project like in Figure~\ref{fig:screen-shots}, and then choose a dashboard that includes a concrete business question $Q$ whose ground-truth answer $A(Q)$ can be directly obtained from the dashboard. Each such pair $(Q, A(Q))$ would then form one benchmark query in \bench.

For example, consider the visualization highlighted by the dashed pink box in the dashboard of Figure~\ref{fig:screen-shots}(a), which corresponds to the business question
$Q =$ \codeq{sales amount, budget, and forecast comparison by month and year}, as indicated by its title. Using the ``Export data'' option shown in Figure~\ref{fig:screen-shots}(b), we can export the underlying result table behind this visualization as a CSV file, which is the ground-truth answer $A(Q)$ for $Q$. 
This $(Q, A(Q))$ pair therefore forms a real test case in \bench.

\bench is the first testbed designed to systematically evaluate LLMs on real-world, end-to-end BI workflows, across all stages of BI. As such, it complements existing data analysis benchmarks, such as NL2SQL, that primarily focus on the analysis step. 

\minihead{\sys: LLM agent for automating end-to-end BI with tools and post-training.}
Leveraging \bench, we study the central research question of whether, given a collection of raw input tables $\mathcal{T}$ and an ad-hoc business question $Q$, LLMs can automatically perform the necessary steps in BI (e.g., select, transform, join, and analyze) to ultimately produce the correct answer $A(Q)$.

We conduct extensive evaluations of state-of-the-art reasoning and chat-based LLMs on \bench. While these models can reason and successfully execute certain BI steps, our results show that they still fail on over 50\% of queries using SQL. In particular, LLMs struggle with operations such as predicting joins and transformations over complex schemas, highlighting the data management challenges that LLMs face and the substantial room for improvement.

Since these challenges have long been studied in the data management community with specialized algorithms, we propose \sys, a reference implementation that integrates data management solutions as ``agentic tools'', which LLMs can invoke and reason over in an agentic tool-call loop (architecture in Figure~\ref{fig:architecture}). 

Observing that LLMs can still struggle with complex multi-step BI workflows in \sys,
we develop a post-training framework that automatically synthesizes training trajectories to post-train the backbone LLM models, which substantially improves models' capabilities for solving end-to-end BI problems through supervised fine-tuning (SFT) and reinforcement learning (RL).


We perform extensive experiments on \bench using \total models and systems.
Figure~\ref{fig:cost-acc-comp} highlights a subset of our results.
We see that (1) \sys's tool design improves accuracy by up to 40 percentage points across frontier LLMs (\textbf{\textcolor{tool_arrow}{teal}} arrows), and (2) \sys's post-training framework yields
significant further gains (\textbf{\textcolor{posttrain_arrow}{purple}} arrows), enabling a small Qwen3-8B to
match or exceed much larger models in quality at up to
50$\times$ lower cost.


\minihead{Contributions.} We summarize our contributions as follows:

\begin{itemize}[leftmargin=*]
        \item We collect and release the first end-to-end BI benchmark, \bench, built using real queries extracted from real BI projects in the wild, to evaluate LLMs' ability on end-to-end BI tasks. 
        \item We conduct extensive experiments using frontier reasoning and chat-based LLMs on \bench, which reveal clear limitations of LLMs on complex data management steps such as joins and transforms, and highlight substantial room for improvement.
        \item We develop \sys, an agentic system that equips LLMs with data management primitives as tools and incorporates domain expertise to address key limitations of general-purpose LLMs in end-to-end BI, yielding substantial accuracy gains. 
        Building on data management primitives from prior work, 
        \sys's novelty lies in the system-level co-design of LLM orchestration and domain-informed tools for BI workflows.
        \item We introduce a post-training framework with a novel trajectory-synthesis method tailored to the BI-domain, enabling SFT and RL to achieve strong improvements on both \bench and out-of-domain data analysis tasks such as Spider 2.0. We show, for the first time, that combining tool use with domain-specific
post-training produces synergistic gains on end-to-end BI tasks. 
    \end{itemize}



%% file: tex/2-Related.tex
\section{Related work}
\label{sec:related}
We review related work in the following areas.

\minihead{Business Intelligence.}
There is a wide range of BI software designed to help users build dashboards and perform ad-hoc data analysis, with Tableau~\cite{tableau} and Power BI~\cite{pbi} being the leading vendors~\cite{gartner-bi}. These platforms provide intuitive visual drag-and-drop interfaces~\cite{mackinlay2007show},  which are popular among non-technical users. However, constructing BI dashboards end-to-end, from raw data to fully functional dashboards, still requires navigating complex BI workflows such as transforming raw data~\cite{m-dsl, tableau-prep} and defining join relationships~\cite{tableau-join, pbi-join}, both of which remain significant pain points for enterprise users~\cite{gartner-bi}. 

\minihead{Data analysis tasks and benchmarks.} While data analysis is well-studied with many established benchmarks, \bench differs fundamentally from prior work in several important ways.

First, while there are many NL2SQL benchmarks (such as BIRD~\cite{bird} and Spider~\cite{spider2}), they focus on the final analysis step, as tables in these benchmarks are already well cleaned, structured, and ready for analysis, which do not capture the full data preparation challenges in real BI workflows as reflected in \bench.

Second, while existing benchmarks span different domains, such as DSBench~\cite{dsbench} and DA-Code~\cite{dacode} in the \textit{data science} domain (with workflows extracted from Jupyter notebooks), KramaBench~\cite{kramabench}, LEAP~\cite{leap}, and REPRO-Bench~\cite{hu-etal-2025-repro} in the \textit{scientific} domain (with workflows extracted from scientific papers), there is currently no benchmark targeting the BI domain. BI workflows present unique challenges, such as (1) messy and semi-structured raw data (e.g., exports from Excel) that require substantial structural transformations; and (2) complex BI schemas that often follow star, snowflake, or constellation designs~\cite{dw-design-1, dw-design-2, dw-design-3} not common in other domains.

Third, while some existing data analysis benchmarks (e.g., InfiAgent-DABench~\cite{infiagent} and SQaLe~\cite{sqale}) rely on synthetically generated queries (e.g., queries generated by LLMs), \bench is constructed entirely from real business questions extracted from real dashboards built by users in the wild, therefore providing a realistic testbed for evaluating LLMs in end-to-end BI workflows.

\minihead{Methods to optimize BI workflows.} There is a long and fruitful line of research in the data management community, addressing \emph{individual} challenges (e.g., join and transform) in BI workflows \cite{zhang2010multi, chen2014fast, hpi, fk-ml, auto-bi,  auto-pipeline, sharma2025datamorpher, ge2025monteprep, tde, flashfill, jin2017foofah, auto-tables, barowy2015flashrelate, jin2020auto, zhu2017auto}.
In this work, we investigate the possibility of leveraging these  algorithms as agentic tools to greatly enhance models' ability in end-to-end BI.

\iftoggle{full}
{

    \minihead{Data analysis agents.} 
    There is a growing body of work on data agents, ranging from early systems such as OpenAI’s Code Interpreter~\cite{code-interpreter} (formerly known as Advanced Data Analysis) and ReAct-based analytical agents~\cite{yao2022react}, to more recent  agentic and post-training methods for tasks like NL2SQL~\cite{bird-leaderboard, spider-2-leaderboard, databao_agent, kaelio2026ktx}. While these systems have achieved strong performance on traditional data analysis tasks, they fall short on end-to-end BI workflows that require complex data manipulations, as we will empirically show in Section \ref{sec:exp}.

}




%% file: tex/3-Problem.tex
\section{Problem: \problem}
\label{sec:problem}
In this section, we begin by introducing the necessary preliminaries, 
followed by a formal definition of our \textit{``end-to-end BI''} problem. 

\begin{figure*}[t]
    \graphicspath{{figures/}}
    \centering

        \includegraphics[width=\linewidth]{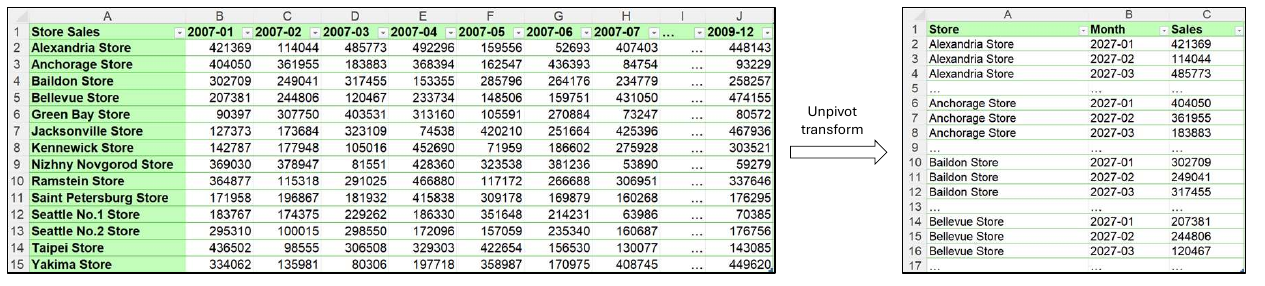}

    \caption{(\underline{Left}): a raw \codeq{Sales}  table imported from spreadsheets. This is a ``pivot table'' in a cross-tabulation form and not properly relationalized, with different months as column headers, which makes it hard to perform relational analysis, such as joining on or aggregating over months. (\underline{Right}): A properly relationalized version of the same \codeq{Sales} table, using the \texttt{unpivot} transformation, where both aggregating over and joining on \codeq{Month} become straightforward.}
    \label{fig:ex-unpivot}
\end{figure*}
\subsection{Preliminaries}
\label{sec:preliminary}
Existing BI workflows require users to perform steps  such as: (1) searching for relevant data, (2) performing data transformations, (3) building join relationships, before (4) dashboards can be built for analysis~\cite{tableau-workflows, pbi-workflows}. We give an overview of these steps below.

\minihead{Search relevant data.}
Upstream data sources (e.g., databases or file stores) often contain numerous files and tables, such that selecting the subset of tables relevant to a specific analysis query is challenging.
For instance, Figure~\ref{fig:screen-shots}(c) shows a small subset of tables from a real BI project. To answer the business question \codeq{Sales and budget by month}, users must inspect and understand available tables in order to locate relevant ones, which poses a substantial burden especially for users unfamiliar with the source data.

\underline{Techniques to automate search.}
Various techniques have been developed in the data management community to automatically rank and retrieve tables for a given user query~\cite{kw-db-1, kw-db-2, kw-db-4}, and more recently LLMs have emerged as strong candidates for table search, given their ability to understand both user natural language queries and tabular data~\cite{llama3, gpt4}.

\minihead{Transform raw input tables.}
Since tables in BI projects originate from heterogeneous sources (e.g., spreadsheets, files, etc.), they are often not analysis-ready and can require diverse transformations.

Broadly, there are two common types of data transformations:
(1) ``\emph{\underline{row-to-row transformations}}''~\cite{tde, flashfill} (e.g., using filtering, string manipulation, arithmetic computations, etc.), which operate on a row-by-row basis; and
(2) ``\emph{\underline{table-reshaping transformations}}''~\cite{auto-tables, barowy2015flashrelate}, which operate at the entire table level, by altering the layout of non-standard and non-relational tables (using operators such as pivot~\cite{op-pivot}, unpivot~\cite{op-stack}, transpose~\cite{op-transpose}, and wide-to-long~\cite{op-wide-to-long}), to produce standard relational tables that are amenable to analysis.

While ``row-to-row transformations'' are relatively straightforward, we review ``table-reshaping transformations''~\cite{auto-tables, barowy2015flashrelate} and explain why they are needed for relational analysis in an example.

\begin{example}
\label{ex:unpivot}
[Table-reshaping transformations]
Figure~\ref{fig:ex-unpivot} (Left) shows a raw \codeq{Sales} table in a BI project, imported from a spreadsheet file, which corresponds to the \codeq{Sales} table in Figure~\ref{fig:screen-shots}(c). This is known as a ``pivot table''~\cite{pivot-table-1, pivot-table-2}, a format commonly used in spreadsheets to organize data into a matrix-like cross-tabulation (e.g., with \codeq{store-names} on the rows and \codeq{month-names} on the columns), enabling users to inspect data values across both row and column directions and identify trends more easily.

While convenient for humans, such pivot tables are known to be ``non-relational''~\cite{auto-tables, db-textbook} and not amenable to relational analysis. In this example, because columns contain homogeneous sales data (which should collapse into the same column), they are not amenable to relational aggregation. For example, to calculate total sales across months, one would need to sum across numerous columns in Figure~\ref{fig:ex-unpivot} (Left). This is in contrast to when the table is properly ``relationalized'' into a table like in Figure~\ref{fig:ex-unpivot} (Right), where a simple range filter is sufficient for the aggregation.

Similarly, joins become challenging when tables are not relationalized.
In order to analyze \codeq{Sales and budget by month} in Figure~\ref{fig:screen-shots}(b), the \codeq{Sales} table in Figure~\ref{fig:ex-unpivot} (Left) must be joined with the \codeq{Budget} table on the \codeq{Month} column. However, the \codeq{Sales} table encodes \codeq{Month} values as column headers, whereas \codeq{Budget} represents them as values in a column, making a join impossible unless the former is properly transformed.

When tables like \codeq{Sales} are not properly structured into relational forms, humans as well as LLMs can struggle to perform relational analysis (e.g., join and aggregation), making these transformations important in end-to-end data analysis.
\end{example}

In addition to unpivot explained in Example~\ref{ex:unpivot}, there are many additional ``table-reshaping transformation'' operators (transpose, pivot, etc.)  needed to transform diverse forms of non-relational tables. These operators are supported in Python Pandas~\cite{python-pandas-api} as well as in proprietary DSLs~\cite{m-dsl} used by BI platforms. In the interest of space, we refer readers to~\cite{auto-tables, barowy2015flashrelate, python-pandas-api} for details of these operators.

Existing BI platforms typically require users to perform these transformations using either vendor-specific DSLs~\cite{m-dsl} or GUI tools~\cite{tableau-prep}, which are clearly challenging for non-technical users.

\underline{Techniques to automate transformations.} Given an analysis question expressed in natural language, we find that LLMs can usually successfully perform the required ``{row-to-row transformations}'' on the fly (e.g., string manipulations or arithmetic computations), by correctly generating the required Python or SQL code.

However, for ``table-reshaping transformations'' (e.g., pivot, unpivot, transpose, wide-to-long), LLMs often struggle both to recognize whether such operations are needed and to execute them correctly, leading to failures in downstream analysis.
 This is likely because (1) LLMs lack a holistic understanding of table structures to identify the need to ``relationalize'' tables; (2) even when LLMs recognize the need to reshape tables, they often struggle to generate correct Python, while SQL lacks native reshaping operators, forcing convoluted workarounds
 \iftoggle{full}
 {
     \footnote{E.g., given the lack of native operators, the workaround for simulating unpivot in Figure~\ref{fig:ex-unpivot} in SQL is to exhaustively union all columns as follows: \\
        {
            SELECT Store\_name, "2007-01" AS Sales, '2007-01' AS Year\_month FROM T \\
            UNION ALL \\
             SELECT Store\_name, "2007-02" AS Sales, '2007-02' AS Year\_month FROM T \\
             UNION ALL \\
             ...
        }
    }
}
that are error-prone.


In the data management literature, specialized algorithms have been developed to automatically predict such reshaping transformations based on the characteristics of input tables~\cite{auto-tables, barowy2015flashrelate, huang2024relationalizing}, which can help LLMs to successfully perform such transformations.


\minihead{Join between tables.}
Building join relationships is another challenge in BI, especially when users are dealing with many tables and complex schemas, like in Figure~\ref{fig:screen-shots}(c). This is amplified by the prevalence of cryptic ID values and surrogate key columns (e.g., store-id, product-id, promotion-id), with similar-looking values from overlapping ranges, leading to false-positive predictions.

\underline{Techniques to automate join.}
There is a long and fruitful line of work in data management on join prediction~\cite{zhang2010multi, chen2014fast, hpi, fk-ml, auto-bi}. In BI workflows, schemas often follow well-known graph structures such as star, snowflake, or constellation. Existing algorithms that exploit these global structural patterns show significantly improved prediction accuracy~\cite{auto-bi}, making them well-suited for BI settings.

We note that while LLMs can be prompted to predict joins, they often underperform because tables in complex BI projects are large (e.g., with millions of rows), forcing prompts to include only a small fraction of rows per table and making it hard to detect reliable value overlap for join prediction. In contrast, classical join prediction algorithms compute statistical features like value containment over full tables~\cite{zhang2010multi, chen2014fast, hpi, fk-ml, auto-bi} and perform inference globally across all tables, making them more reliable choices when dealing with large tables and complex schemas.

\minihead{Data analysis for answering business questions.} Finally, after relevant tables are identified, transformed, and joined, existing BI platforms allow users to use drag-and-drop UI to build dashboards as shown in Figure~\ref{fig:screen-shots}(a), where users can specify filtering, grouping, aggregation, as well as write custom calculation logic to produce desired results that are visualized on dashboards. Although such analysis is expressed through vendor-specific interfaces, the underlying logic can be equivalently mapped to executable SQL or Python code, which is the focus of our research.

\underline{Techniques to automate analysis.}
General-purpose LLMs have become promising in generating SQL or Python snippets that can perform ad-hoc analysis~\cite{gpt4, llama3}, as reflected in the significant progress on tasks such as NL2SQL~\cite{bird-leaderboard, bird, spider-2-leaderboard, spider2}, making LLMs strong candidates to automate the final analysis step, to produce final answers to users' analytical questions.


\subsection{Problem Definition}
\label{subsec:problem}

Given the end-to-end BI workflows introduced in the previous section, we now define our ``\problem'' problem  as follows.

\begin{definition}
\label{def:problem}
[\problem] Given a natural language query $q$ and a collection of raw input tables $\mathcal{T} = \{T_1, \dots, T_n\}$, the \textit{``end-to-end BI''} problem requires a system to automatically perform necessary search, transform, join, and analysis steps to produce a final result table $R$ that can correctly answer $q$ on $\mathcal{T}$.
\end{definition}

\begin{example}
\label{ex:problem}
[\problem] Given an ad-hoc question $q =$ \codeq{Show sales amount, budget, and forecast comparison by month and year} shown in the visualization title of Figure~\ref{fig:screen-shots}(a), in end-to-end BI, a system would perform all necessary steps in BI workflows automatically: (1) \emph{search} to identify relevant tables such as \codeq{Sales}, \codeq{Budget}, and \codeq{Date}, (2) \emph{transform} to perform transformations such as unpivot on the \codeq{Sales} table as shown in Figure~\ref{fig:ex-unpivot} and Example~\ref{ex:unpivot}, (3) \emph{join} to establish relationships across tables like in Figure~\ref{fig:screen-shots}(c), and finally (4) \emph{analysis} to produce the final result table $R$ like in Figure~\ref{fig:screen-shots}(b), which can correctly answer the given query $q$.
\end{example}

\textbf{Relation to traditional BI.} We note that the research problem we propose in Definition~\ref{def:problem} targets a regime complementary to traditional BI. While traditional BI assumes data has been cleaned/transformed through ETL, and organized into a curated data model; we consider one-off, low-touch analysis over freshly acquired or external raw tables, where such a data foundation does not yet exist and users are unable or unwilling to incur the substantial up-front cost of constructing it merely to answer a small number of questions. Our work asks whether LLMs can enable useful analysis for this emerging class of long-tail users; and we do not intend to position it as a full replacement of mature enterprise BI systems.

We also note that while BI analysis is traditionally rendered as dashboard visualizations, there is no unique ground truth for what visualization a result table $R$ should use (as it is possible to visualize $R$ differently~\cite{stolte2002polaris, grammar-of-graphics}). Therefore, in this work, we focus on producing the correct result table $R$ for the given query $q$ and data $\mathcal{T}$, without requiring an LLM to produce a visualization exactly matching the user-authored visualization in real BI projects.

Thus, in this
paper, \textit{``end-to-end'' refers to automating the complete raw-table-to-answer
workflow defined above}, 
rather than the full 
lifecycle of an enterprise BI
deployment.





%% file: tex/4-Data.tex
\section{\bench: End-to-end BI Benchmark}
\label{sec:benchmark}

Since existing benchmarks like NL2SQL focus primarily on the data analysis stage, we construct a new benchmark, \bench, to better reflect the end-to-end challenges in real-world BI.


\minihead{Dataset Preprocessing.}\label{subsec:data_prep}
We crawl over 3K real Power BI project files (.pbix files) hosted at publicly accessible URLs using a web search engine. Each project file is self-contained  with raw data files, user-programmed transformations and joins, to ultimately build dashboards to answer business questions. Many of these dashboards reflect substantial sophistication and effort, as shown in Figure~\ref{fig:screen-shots}.

From each BI project, we programmatically extract its raw data files and metadata, including (1) user-generated transformation steps, (2) manually defined join relationships, and (3) the final analytical dashboards, using DAX-studio and RPA (power-automate), following a process similar to that described in~\cite{auto-bi}. These artifacts capture real user-created end-to-end workflows and serve as the basis for our study. 



\minihead{Test Case Generation and Validation.}\label{subsec:data_label}
We spend substantial manual effort curating and verifying a total of 100 pairs of (analytical query $q$, ground-truth table $R$) from sampled real BI projects, 
making sure that each ($q$, $R$) pair is carefully reviewed and validated in a systematic process. We document this  process below: 




\underline{Test case collection.}
We repeat the following sample-screen-verify procedure until we reach a cap of 100 test cases\footnote{\bench with 100 instances is on
par with the most closely related benchmark efforts for agentic data science tasks: e.g., REPRO-Bench (112
tasks)~\cite{hu-etal-2025-repro}, KramaBench (104
tasks)~\cite{kramabench}, DS-Agent (30 tasks)~\cite{guo2024dsagentautomateddatascience}, and TAG (80 tasks)~\cite{biswal2024text2sqlenoughunifyingai}, which are all designed to provide a focused evaluation of a specific task at a reasonable labeling cost.}, a target that balances the substantial manual curation effort required (constructing \bench took over 400 person-hours, or over 4 person-hours per query), and statistical reliability (we find 100 test cases provide strong statistical power in our proposed evaluation, as we will report in our experiments). In each iteration,
we first randomly sample a visualization from user-created dashboards in a BI project, and select only visualizations that satisfy the following conditions as the basis for generating a test query $q$: \textit{(1)} The intent of the visualization is semantically clear based on its context (e.g., its title and surrounding annotations), so that we can paraphrase the intent into a natural language query (when the meaning of the visualization is ambiguous due to short or cryptic titles, we discard such cases);
\textit{(2)} The query paraphrased from the visualization title and surrounding context accurately reflects the analysis (we verify this by inspecting the computation logic behind the visualization);
\textit{(3)} The data and the visualization are in English (as otherwise we could not verify 
their semantic correctness).

Similar to existing NL2SQL and agentic benchmarks~\cite{bird,spider2}, where annotators formulate natural-language questions based on user intent, the annotators for \bench observe real user's query intent from existing visualizations (from user-authored title and surrounding context), paraphrase such intent in a natural language query $q$, and verify that the query is consistent with the underlying computation logic. 
In this way, each test query $q$ captures real users' query intent reflected in real artifacts, and therefore serves as realistic tests in \bench.
Figure~\ref{fig:screen-shots}(a) shows such an example: the visualization highlighted in the dashed pink box is titled \codeq{Sales amount, budget and forecast comparison by month and year}, which accurately describes the analysis that we can paraphrase into a query (e.g., \codeq{Show me sales amount, budget and forecast comparison by month and year}) in our benchmark.


\iftoggle{full}
{
    In some cases, the visualization title is only a simplified version of the underlying analytical query and may omit implicit filters that are applied in the actual visualization. We manually inspect the visualization DSL logic and its outputs, and revise the query to include any missing filter conditions. For instance, if we determine that a filter \codeq{year >= 2007} is applied to produce the result in Figure~\ref{fig:screen-shots}(b), we update $q$ to \codeq{Sales amount, budget, and forecast comparison by month and year \underline{since 2007}}.
}

For each verified $q$, we use the ``export'' functionality to extract the result table $R$ corresponding to the visualization, as shown in Figure~\ref{fig:screen-shots}(b), which produces candidate query-result pairs: $(q, R)$.

\underline{Ground-truth verification.}
We perform extensive verification to ensure the quality of each test case, making sure that there are no ambiguous queries, missed data filters, or issues in our extraction steps that may lead to mismatches between queries and results.

In addition to manual review and verification, we also perform an LLM-assisted verification step, by providing two independent frontier LLMs 
with the query $q$, along with all types of simplifying ``hints'' (e.g., necessary transformations and transformed tables, ground-truth joins, and the subset of tables actually relevant to $q$, etc.). If the models fail to reproduce the expected result table $R$ even once after 10 attempts in this simplified setting (which happened in
23.1\% of cases), it triggers another round of extensive manual review and verification -- specifically, we inspect the query, the visualization's computation logic, and our extraction steps; if we identify an error in our extraction process, we repair the $(q, R)$ pair (14.4\% of all cases); or if manual review confirms genuine semantic mismatch or ambiguity  we discard such cases (5.2\% of all cases). Importantly,  when
manual review confirms that the $(q,R)$ pair is correct and unambiguous, cases that neither model can solve are retained unchanged (3.5\% of
all cases).

Thus, model success is not an inclusion criterion: the LLM-assisted check
serves only to trigger another round of extensive human reviews (requiring 1.5 hours per case on average), rather than a convenient filter by model solvability.

\underline{Ground-truth augmentation}. Prior research on NL2SQL shows that a unique ground-truth table can be inadequate, as LLM-generated outputs may deviate slightly in schema from the ground truth yet remain semantically correct to human users~\cite{wretblad2024understanding, floratou2024nl2sql}.

For instance, for the query \codeq{Difference between income and expenses of all departments}, the ground truth may be labeled to have two columns, \codeq{department} and \codeq{difference}, whereas an LLM may produce a table that additionally includes columns \codeq{income} and \codeq{expenses}. While the inclusion of the two additional columns in the result is useful, a simple automated evaluation can nevertheless mark the result as incorrect due to a schema mismatch. (Conversely, if the ground truth includes all four columns but the LLM returns two, that can also be marked as incorrect.)

To address this issue, when building \bench, we carefully augment each test case with multiple semantically valid ground-truth tables. In the example above, if the ground-truth result table $R$ extracted from a user visualization contains two columns, \codeq{department} and \codeq{difference}, but after review we determine that including \codeq{income} and \codeq{expenses} is also semantically valid (since they are also referenced in the query), we augment the test case with an alternative ground truth containing all four columns. All such augmentations are based solely on the query semantics and the visualization's computation logic (no model output is consulted), and are reviewed and agreed upon by two domain experts to ensure consistency and quality.

After this rigorous curation process, the resulting 100 test cases, each consisting of an analytical query $q$ and a ground-truth table $R$, form our \bench benchmark.


\minihead{Benchmark Analysis.}
\label{subsec:data_stat}
Our analysis of \bench shows that on average, each project contains 10.6 tables (the maximum is 52), 80,464 rows (the maximum is over 7M), and 10.7 columns per table (the maximum is 325). There are on average 12.75 join relationships in each project (the maximum is 94).
We highlight two additional observations from our benchmark analysis below.


\begin{figure}[t]
    \graphicspath{{figures/}}
    \centering
    \includegraphics[width=0.65\columnwidth]{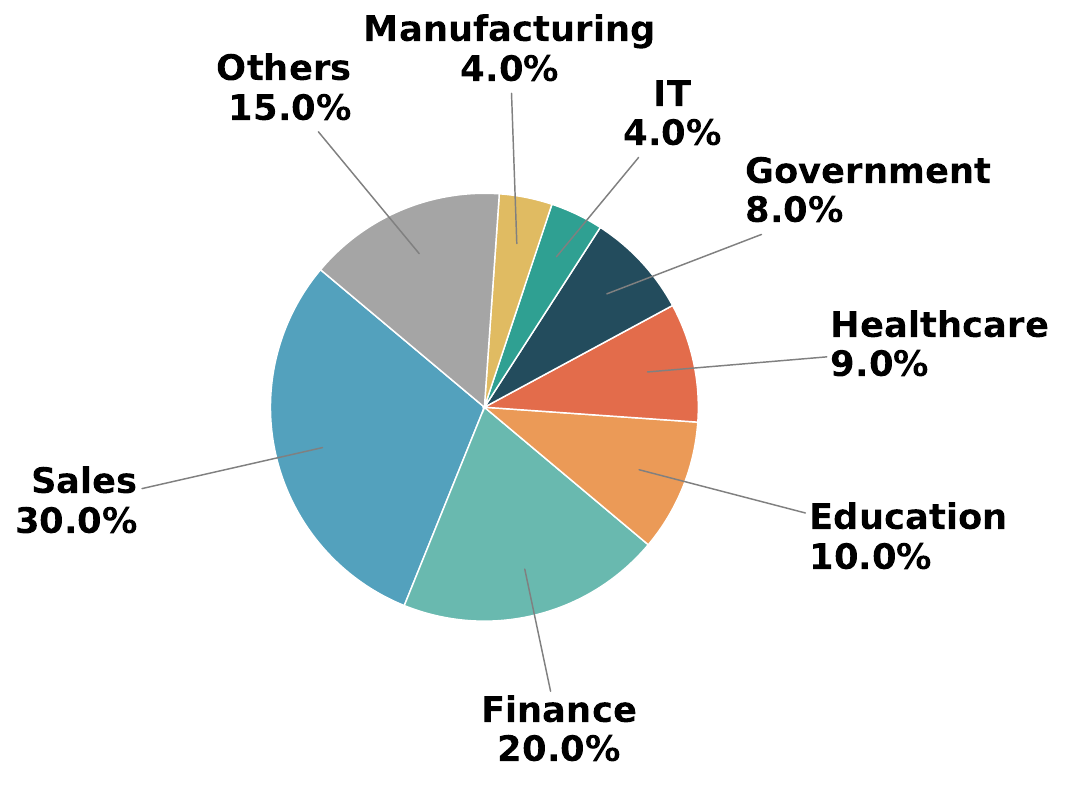}
    \caption{\bench: distribution by domains.}
    \label{fig:bench_domains}%
\end{figure}

\underline{\bench is challenging.}
We find that \bench is highly challenging: even frontier models consistently fail on over 50\% of the test cases using SQL.
For example, the best-performing model, o4-mini, achieves only 48.2\% accuracy using SQL (and it frequently requires many rounds of interaction). This is not surprising: one test query in \bench requires models to reason over 38 tables (all of which are required to answer the query), making table selection, transformation, and join all challenging.

Similarly, state-of-the-art NL2SQL systems struggle on \bench: the top four open-source models on the BIRD leaderboard
and the top two open-source agents on the Spider~2.0-lite
leaderboard all score above 70\% on their respective benchmarks,
yet achieve only 6.0--17.3\% and 23.8--26.3\% SQL accuracy on
\bench, respectively (see Table \ref{tab:main_res_post_train} for details).
This gap arises because \bench captures end-to-end data challenges in BI workflows, extending beyond pure data analysis (which is the primary focus of NL2SQL systems).



\underline{\bench is diverse.}
We classify the business domains of \bench tasks in Figure \ref{fig:bench_domains}. \bench covers diverse domains, ranging from sales, finance, and education to areas such as sports and transportation (grouped in the ``other'' category). The diversity of \bench reflects the broad range of analyses performed by BI users in practice, which also contributes to its difficulty.

%% file: tex/5-Method.tex
\section{\sys: AI agent for End-to-End BI}
\label{sec:agent}

We now introduce \sys, an agentic framework we develop for end-to-end BI that combines both tool use and post-training. \sys is a unified design that integrates LLM reasoning with traditional data management methods and improves LLMs' ability to orchestrate them effectively.


\begin{figure*}[t!]
    \graphicspath{{figures/}}
    \centering
    \includegraphics[width=0.9\textwidth]{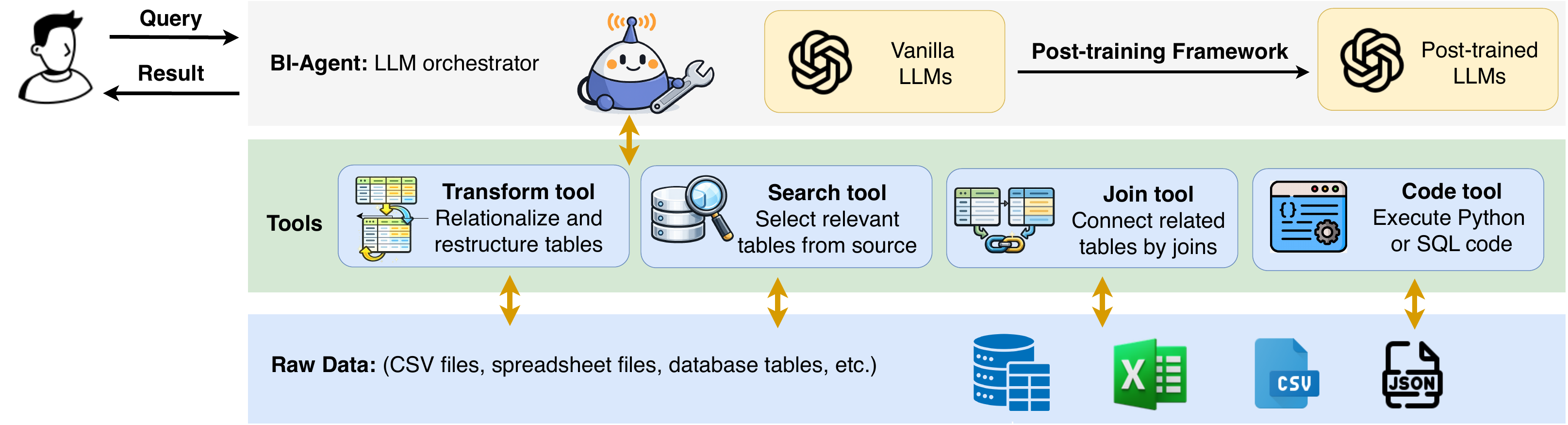}
    \caption{Architecture overview of \sys. The top layer shows \sys uses different LLMs (vanilla vs. post-trained) as the orchestrator, to plan and execute a reasoning/tool-call loop. The middle layer shows tools such as general coding and data-management tools that can be invoked by \sys, which \sys uses to interact with raw data files in the bottom.
    }
    \label{fig:architecture}%
\end{figure*}

\subsection{Architecture: reasoning and tool-call loop}
\label{subsec:tools}

Recent LLMs can iteratively reason and interact with external tools~\cite{react, toolformer}, making incremental progress toward solving a task. These capabilities are now natively supported in mainstream models, which can take a set of tool descriptions as input and autonomously decide when and which tools to invoke~\cite{function-calling-anthropic, function-calling-oai}.

Observing the data management challenges in end-to-end BI (Section~\ref{sec:preliminary}), in this work, we design a \sys framework that leverages LLM's reasoning and tool-call loop, where we expose algorithmic and LLM-based data management methods as agentic tools, to better solve the hard data challenges in end-to-end BI tasks.

Figure~\ref{fig:architecture} shows the overall architecture of \sys.  At the top layer, we have the agent framework that can be instantiated using different LLMs as \textit{orchestrators}, which can be both vanilla LLMs and post-trained LLMs for BI tasks (to be discussed in Section \ref{subsec:training}). In the middle, we have a collection of agentic tools leveraging specialized data management methods (e.g., \texttt{transform}, \texttt{search}, \texttt{join}), together with a generic \texttt{coding} tool, that LLMs can invoke in their reasoning and tool-call loop, which we will discuss next. These tools, in turn, interact with raw data imported from places like databases or CSV files, shown at the bottom of the figure.



\minihead{Tool design.} We now describe the tools used in \sys, including a coding tool common in agent systems, along with three specialized data management tools: \code{transform}, \code{join}, and \code{search}.

\underline{Coding tool}.
Code serves as a common interface through which models interact with the environment in many agent designs (e.g., NL2SQL-style tasks). Similarly, we implement a standard coding tool with a runtime environment that executes LLM-generated code (e.g., Python or SQL) and returns execution results or error messages as feedback. Program variables from previous interaction rounds are cached and remain accessible, enabling LLMs to make iterative progress toward solving complex BI tasks.

\underline{Transform tool}. Transform is the first specialized data management tool we introduce in \sys. As discussed in Section~\ref{sec:preliminary}, complex table-reshaping transformations are required to convert diverse raw tables into relational forms to facilitate relational analysis, yet LLMs struggle with this task because they lack a holistic understanding of table structures to identify the need to ``relationalize'' tables,
and even when they recognize the need, they often fail to synthesize reshaping steps correctly (especially when such steps are not natively supported  as standalone operators, e.g., in SQL). It is therefore beneficial to invoke transformation prediction as an external tool to offload this challenging data step from LLMs.

We build a transformation tool using prediction algorithms developed in the data management literature, specifically optimized for table-reshaping operations~\cite{auto-tables}.
This tool inspects all input raw tables and determines if reshaping steps such as transpose, pivot, and unpivot are needed. It applies the predicted transformations and returns both the original and transformed tables as output. The LLM at the orchestration layer can then take the output from this tool for additional processing (e.g., for join or relational analysis).

\underline{Join tool}.
Identifying joins accurately is a prerequisite for complex analysis over multiple tables, yet LLMs frequently make join mistakes over complex schemas (Section~\ref{sec:preliminary}).
We introduce a specialized join tool by implementing a join prediction algorithm from the data management literature~\cite{auto-bi}, which is specifically optimized for snowflake-like schemas common in BI settings. This tool takes a list of input tables and returns all predicted joins as output, so that the LLM orchestrator in \sys can leverage the predicted joins to perform analysis over complex schemas more effectively.


\underline{Search tool}. Search is another specialized tool in \sys.
When operating over large collections of tables, LLMs must spread attention across many tables and columns at the same time, making them prone to confusion from semantically related but irrelevant tables, which often leads to failed analyses.

Similar to human analysts, restricting attention to tables relevant to a target query $q$ improves the likelihood of correct analysis. To this end, we implement a specialized search tool specifically focused on the task of selecting tables relevant to the given $q$. Given $q$ and available tables (represented by column headers and sampled rows), the tool prompts an LLM to identify a subset of potentially relevant tables, using a conservative strategy that retains any plausibly relevant tables while pruning away those that are clearly irrelevant.

\underline{Extensibility}. Note that the tools we design in \sys are not meant to be exhaustive, and can be easily extended to incorporate additional data management primitives in the same LLM tool-call loop, to further enhance LLM-driven end-to-end BI.

In \sys, we provide LLMs with descriptions of all available tools, and instruct them to invoke these tools as appropriate.
We use an example below to illustrate how \sys works with tools.

\begin{figure}[h!]
\centering
\begin{adjustbox}{max width=1\linewidth}
\begin{minipage}{1.1\linewidth}
\begin{lstlisting}[
basicstyle=\ttfamily\small,
breaklines=true,
numbers=left,
numberstyle=\scriptsize,
caption={Example tool-call trajectory},
label={lst:trajectory},
columns=fullflexible,
xleftmargin=1.5em
]
{"tool": "transform_table", "input": "..."},
{"tool_response": {"output": "tables: [(Sales, unpivot)]"}},
{"tool": "search_tables", "input": "..."},
{"tool_response": {"output": ["Sales", "Budget", ...]}},
{"tool": "join_tables", "input": ["Sales", "Budget", ...]},
{"tool_response": {"output": "(Sales.C1, Budget.C2)"}},
{"tool": "execute_code", "input": "import pandas as pd..."},
{"tool_response": {"output": <execution results>}},
...
\end{lstlisting}
\end{minipage}
\end{adjustbox}
\end{figure}

\begin{example}
\label{ex:e2e}
We revisit Example~\ref{ex:problem}. Recall that the user query is $q =$ \codeq{Sales amount, budget, and forecast comparison by month and year}, taken from Figure~\ref{fig:screen-shots}(a). 
Listing~\ref{lst:trajectory} shows an example LLM trajectory to solve $q$, which we explain in detail below.

In Line 1, the LLM orchestrator in \sys first invokes the transform tool. This tool checks the input tables and predicts that the \codeq{Sales} table requires an \code{unpivot} transform. In the next \code{tool\_response} line (Line 2), the tool output (including the prediction and the transformed  \codeq{Sales}) is appended to the model context, to inform the LLM orchestrator in future iterations.

In the next iteration (Line 3), \sys invokes the search tool, which returns a list of tables relevant to $q$, such as \codeq{Sales}, \codeq{Budget}, etc. The tool output is also appended to the model context (in Line 4) to inform future model iterations.

\sys then invokes the join tool (Line 5), which adds a list of joinable tables and columns in the model context too (Line 6).

In the next iteration, \sys invokes the coding tool, which uses the relevant information from previous iterations and starts performing analysis. The result of the code execution is appended in Line 8. This process continues iteratively until the LLM orchestrator determines that it has produced an answer to the given question $q$, at which point it stops and exits the reasoning and tool-call loop.
\end{example}

\begin{figure*}[t]
    \graphicspath{{figures/}}
    \centering
    \includegraphics[width=0.9\textwidth]{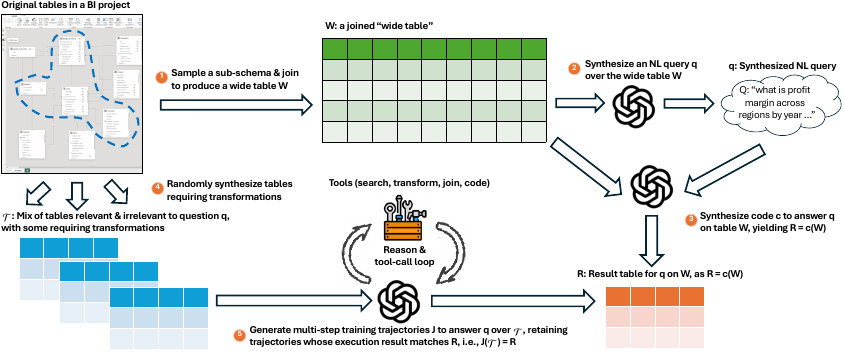}
    \caption{Data generation pipeline to automatically synthesize training trajectories for model post-training.}
    \label{fig:ft_data_collection}%
\end{figure*}

% ...

By combining the LLM orchestrator's reasoning and tool-calling loop with specialized data management tools, difficult data challenges can be offloaded from LLMs, significantly improving overall model success rates, as we will show in our experiments.

\subsection{Post-training models for end-to-end BI}
\label{subsec:training}

While \sys significantly improves LLMs' ability to perform end-to-end BI tasks (by over 10 percentage points on average, as we will see in Section \ref{sec:exp}), we observe that vanilla models can still fall short in their abilities to (1) use relevant tools when needed, and (2) perform complex coding tasks on tables for analysis.
This motivates us to explore post-training, where we adapt open-source models to better support end-to-end BI scenarios.

\minihead{Synthesize training trajectories.}
While we built \bench by manually curating 100 test cases from real BI dashboards, we could not follow the same steps to generate training data given the significant amount of manual work involved.
Therefore,  we propose a post-training framework with a new
method specifically designed for the BI domain, to systematically and automatically synthesize training tasks tailored to the domain. 

Recall that we collected a large number of real BI projects in the wild (Section~\ref{sec:benchmark}), which we leverage to synthesize  training tasks. 
To prevent data leakage, we enforce both project-level and data-level separation by excluding from training all projects included in \bench, as well as any project with partial table-content overlap (at least 5 rows) with a test project in \bench. (We additionally perform query-level and transform-level leak prevention in our synthesis, which we will discuss later in this section).

We systematically synthesize BI queries and steps like shown in Figure~\ref{fig:ft_data_collection}, which we describe below.
In the first step
\tikz[baseline=(char.base)]{
  \node[shape=circle, fill=orange, text=white, font=\small, inner sep=1pt] (char) {1};
}
of the figure, for each collected real BI project, we use data tables and pairwise join relationships programmatically extracted from the project, to construct a join graph as shown at the top-left of the figure. We then start from a random table and follow
all its join edges to sample a connected sub-schema, as illustrated by the blue circle in the figure. Because we know exactly how these tables should join (from user-programmed joins), we join all of them together to produce a single ``wide table'' $W$, shown in the middle of the figure, which connects diverse information from multiple source tables (e.g., it could connect \codeq{Sales} with \codeq{Budget} on the \codeq{Date} dimension, as in Figure~\ref{fig:screen-shots}(b)).

In step
\tikz[baseline=(char.base)]{
  \node[shape=circle, fill=orange, text=white, font=\small, inner sep=1pt] (char) {2};
}, we prompt an LLM to synthesize a natural language query $q$ on this wide table $W$, and then in step
\tikz[baseline=(char.base)]{
  \node[shape=circle, fill=orange, text=white, font=\small, inner sep=1pt] (char) {3};
}, a corresponding code snippet $c$ that can be used to answer $q$ on the table $W$. We execute $c$ on $W$ to get $R=c(W)$, which we treat as the result table that can answer query $q$. If $R \neq \emptyset$, we retain \(q\); otherwise, we discard it. This process yields 7,985 synthesized queries.

By selecting a sub-schema and constructing a wide table in the first step, we greatly simplify the code generation task. The LLM no longer needs to search for relevant tables from a large collection of raw input tables $\mathcal{T}$ because a selected sub-schema is provided. Moreover, materializing all joins into a single wide table eliminates the challenge of predicting joins over complex schemas, making it much easier to generate the correct code $c$ for question $q$ on $W$.

Next, in step
\tikz[baseline=(char.base)]{
  \node[shape=circle, fill=orange, text=white, font=\small, inner sep=1pt] (char) {4};
}, we turn to the tables extracted from the BI projects and construct a new collection of tables by mixing tables selected in the sub-schema, shown inside the blue circle, that are relevant to query $q$ with tables outside the blue circle that are irrelevant to $q$. We also apply inverse table-reshaping operators to input tables with a small probability, following~\cite{auto-tables}, to generate raw non-relational tables requiring reshaping transformations. This creates complex training cases requiring search and transformations, thereby exercising the model's ability in these areas. 

Finally, in  step
\tikz[baseline=(char.base)]{
  \node[shape=circle, fill=orange, text=white, font=\small, inner sep=1pt] (char) {5};
}, we are ready to generate training data because, at this point, we have synthesized one training task $t = (q, \mathcal{T}, R)$, where the raw input tables $\mathcal{T}$ (in blue) are shown on the left of the figure and the synthesized result table $R$ for query $q$ is shown on the right. In this final step, we generate training trajectories that can correctly go from $\mathcal{T}$ to the target result $R$.


\underline{Leakage prevention in training data.}
Beyond project- and table-level deduplication to prevent leakage as discussed earlier, since our training
tasks are synthesized from scratch by LLMs, they differ substantially from the test tasks collected from real user-authored dashboards in \bench, in both (1) the analytical queries and (2) the transformation required in each task, which naturally prevents 
leakage and overfitting.


To quantify the difference between train/test tasks at both the query- and transform-level, we perform the following analysis. 


At the \textit{query-level}, let $\mathrm{sim}(x,y)$ be cosine similarity of two
queries $x$ and $y$ under \texttt{text-embedding-3-large}~\cite{openai_embedding_models}.
For each test query $x$, we compare its semantic nearest neighbor test query in
\bench, $s(x)=\max_{y\in\mathcal{Q}_{\bench}\setminus\{x\}}\mathrm{sim}(x,y)$, with
its semantic nearest neighbor in the training set, by measuring the fraction of
training queries that are less similar to $x$ than $s(x)$:
$
\rho(x)=\frac{\bigl|\{t\in\mathcal{Q}_{\text{train}}:\mathrm{sim}(x,t)<s(x)\}\bigr|}
{|\mathcal{Q}_{\text{train}}|}.
$
Averaged over the entire query set, $\rho=99.72\%$, i.e. only $0.28\%$ of training queries are more similar to a
test query than that test query's nearest neighbor in \bench, showing strong
dissimilarity between real test queries and synthesized training queries.

At the \emph{transform-level}, repeating this analysis  with $\mathrm{sim}$ replaced by
Jaccard overlap of column-header tokens and each test transform compared against all
training transforms, 
only $0.73\%$ of training transforms are more similar to a test transform than to that test transform's nearest neighbor in \bench, again showing strong separation between train/test transforms.

This separation is further corroborated by the strong generalizability we report
in Section~\ref{subsec:data_ablation}: the gains
from post-training transfer to out-of-domain tasks like NL2SQL that are beyond \bench.


\textbf{Training procedures.}
For \underline{supervised fine-tuning (SFT)}, we use a frontier LLM (GPT-4o) as a ``teacher model'' $M_T$. 
Given the query \(q\) and raw input tables \(\mathcal{T}\), without access to either the synthesis code \(c\) or target result \(R\), \(M_T\) generates an end-to-end trajectory \(J\). We retain \(J\) only if executing it on \(\mathcal{T}\) exactly reproduces \(R\), and then
split it into turn-level SFT examples.
Thus, models are post-trained only on tasks that pass this independent 
semantic verification.

We follow this process and generate thousands of $(t, J)$ training pairs at scale for SFT. Note that this generation framework is generalizable and can be configured to use (1) Python vs. SQL and (2) specialized data management tools vs. no such tools. Table~\ref{tab:data_stats} shows the training trajectories generated for each of the four settings: \{tool, no-tool\} × \{Python, SQL\}, all of which we test in our experiments to validate the effectiveness of model post-training.

\iftoggle{full}
{
    From each successful multi-round interaction trace, we extract training examples at the assistant-turn level. Concretely, for every assistant response, we construct one SFT data point where:
    (1) the input consists of all preceding messages in the conversation history, and
    (2) the target output is the corresponding LLM response.

    To standardize training format, we encode tool specifications in the system message and convert tool calls into structured JSON blocks embedded within the assistant message.
}

In the case of \underline{reinforcement learning (RL)}, we use the Reinforcement Learning with Verifiable Reward (RLVR) paradigm~\cite{shao2024deepseekmathpushinglimitsmathematical, rlvr-1} and the same synthesized analysis tasks $t = (q, \mathcal{T}, R)$ constructed from steps 1-4 of Figure~\ref{fig:ft_data_collection}.
However, instead of using a teacher model $M_T$ to generate trajectories $J$ for a student model $M_S$ to learn from (step 5), we use $R$ as the ``verifiable reward'' and have $M_S$ ``roll out'' and explore different possible trajectories until it produces a successful trajectory whose result matches the target table $R$.

We use the GRPO algorithm~\cite{shao2024deepseekmathpushinglimitsmathematical} popularized by DeepSeek-R1~\cite{r1}. For the same task $t$, we sample $K$ different roll-out trajectories from a current LLM policy $\theta$:
$ J_1, J_2, \ldots, J_K \sim \pi_\theta(\cdot \mid t),$
each invoking relevant tools towards the desired analysis goal.
We calculate the reward of each trajectory $J_i$, denoted as $r_i(J_i | t)$, by comparing the resulting table $\hat{R} = J_i(\mathcal{T})$ against the target table 
$R$:
\begin{equation}
\label{eq:reward}
r(J_i | t) =
\begin{cases}
+1 - 0.1 n, & \text{if } \hat{R} = R \\
-0.5 - 0.1 n, & \text{if } \hat{R} \neq \emptyset \ \text{and } \hat{R} \neq R \\
-1 - 0.1 n, & \text{otherwise}
\end{cases}
\end{equation}
This reward design is motivated by prior RL research as well as our own empirical ablations. Specifically, we introduce the term $-0.1n$ as a format penalty in Equation~\eqref{eq:reward},  where $n$ denotes the number of code-generation iterations that produce syntax errors. This penalty provides an explicit reward signal that encourages the small model to pay attention to the syntactic validity of its code, which is similar to the format penalty used in DeepSeek-V3.2~\cite{deepseekv32}.

In the second line of Equation~\eqref{eq:reward}, when a trajectory correctly follows the task instruction to successfully return a result table $\hat{R}$, which, however, is incorrect, we assign a partial credit (-0.5).
This partial credit is intended to provide a \emph{dense} learning signal in an otherwise sparse-reward setting similar to prior work~\cite{r1} (as most early rollouts fail outright), which turns out to be important as we will show empirically. 
Finally, if the trajectory fails to even return a result table, we give a reward of -1 (the last line of Equation~\eqref{eq:reward}).  

In GRPO, these reward values are normalized into an ``advantage'' score $A_i$, which is calculated for each trajectory $J_i$ as its z-score within the batch 
$A_i$ = $\frac{r_i - \mu}{\sigma}$,
where $\mu$ and $\sigma$ are the mean and standard deviation of the rewards in the same batch ($r_1, \ldots, r_K$). Intuitively, each $A_i$  tells the relative goodness of the trajectory $J_i$ in the same batch.
We then optimize the policy using GRPO~\cite{shao2024deepseekmathpushinglimitsmathematical}:
$
\mathcal{L}(\theta) = \mathbb{E}_{x, \{y_i\}}\left[
\min\left(
r(\theta) A,\
\mathrm{clip}(r(\theta), 1-\epsilon, 1+\epsilon) A
\right)
\right],
$
where $r(\theta) = \frac{\pi_\theta(y \mid x)}{\pi_{\text{old}}(y \mid x)}$.
This objective encourages increasing the probability of outputs with positive advantage \(A\), while preventing 
overly 
large updates via clipping.

During roll-outs, the model is provided with the full conversational context, including intermediate reasoning steps. However, when computing policy gradients, we restrict optimization to the tool-call blocks only (coding, transform, join, etc.), so that RL focuses on learning core tool-calling and coding 
skills rather than mimicking the generated natural language.

Since RL is most effective when training includes both positive and negative outcomes, and positive outcomes are difficult to obtain given the challenging nature of end-to-end BI, we use synthesized analysis tasks $t = (q, \mathcal{T}, R)$ for which GPT-4o is able to generate correct trajectories during the SFT stage as rollout tasks for RL. We report our training data statistics in Table~\ref{tab:data_stats}.

\begin{table}[t]
\centering
\caption{Training data statistics.}
\label{tab:data_stats}
\scalebox{0.75}{
\begin{tabular}{lcccc}
\toprule
& Python No Tool & Python Tool & SQL No Tool & SQL Tool \\
\midrule
\#SFT Turns & 19,563 & 28,698 & 23,100 & 42,686 \\
\#RL Tasks  & 2,975  & 4,586  & 2,049  & 3,426  \\
\bottomrule
\end{tabular}}
\end{table}

%% file: tex/6-experiment.tex
\section{Experiments}
\label{sec:exp}


\subsection{Experiment Setup}
\label{subsec:setup}

\minihead{Models and methods compared.}
We conduct extensive experiments on a total of \total models and systems, including proprietary and open-source LLMs, as well as state-of-the-art NL2SQL methods:
\begin{itemize}[leftmargin=*]
\item For large proprietary models, we evaluate:
\underline{\textbf{(1)}} GPT-5.5~\cite{openai_gpt5p5_api}, a frontier model; \underline{\textbf{(2)}} GPT-5.2~\cite{openai_gpt5p2_chat_api}, a strong general-purpose model;
\underline{\textbf{(3)}} GPT-4o~\cite{openai_gpt4o_api}, a fast and versatile model; and
\underline{\textbf{(4)}} o4-mini~\cite{openai_o4-mini}, a reasoning-optimized model. 

\item For open-source models, we evaluate \underline{\textbf{(5)}} Llama-4-Maverick~\cite{huggingface_llama4maverick17b128e}, a large 400B MoE model; \underline{\textbf{(6)}}
GPT-OSS-120B~\cite{openai2025gptoss}, OpenAI's open-weight reasoning model;
\underline{\textbf{(7)}} DeepSeek-V4-Pro~\cite{deepseekai2026v4}, a 1.6T-parameter
MoE model that is state-of-the-art among open-weight models on
agentic coding benchmarks; \underline{\textbf{(8)}}
Mistral-Large-3~\cite{mistralai2025mistral3}, a 675B-parameter MoE model; \underline{\textbf{(9)}} Kimi-K2.6 \cite{moonshotai2026kimik26}, a 1T-parameter
MoE model optimized for long-horizon agentic coding; and
\underline{\textbf{(10)}} Qwen3-8B~\cite{qwen3technicalreport}, a competitive 8B reasoning model.

\item For post-trained models, we test a total of 8 post-trained model variants summarized in Table \ref{tab:exp_settings}: (SFT vs. SFT+RL\footnote{RL models are initialized from SFT checkpoints; thus we abbreviate SFT+RL as RL.}) $\times$ (No tools vs. Tools) $\times$  (SQL vs. Python), all based on Qwen3-8B (models \underline{\textbf{(11)}}-\underline{\textbf{(18)}}). 
Recall that our post-training uses synthesized training data held completely separate from \bench, to prevent leakage (Section~\ref{subsec:training}).

\item For comparisons with NL2SQL systems, we test against the
best-performing NL2SQL models: \underline{\textbf{(19)}}
Kwai-AutoSQL-14B~\cite{kuaishou2026kwaiautosql14b}, \underline{\textbf{(20)}}
Kwai-AutoSQL-32B~\cite{kuaishou2026kwaiautosql32b}, \underline{\textbf{(21)}}
Infly-RL-SQL-32B~\cite{infly_inf_rl_qwen_coder_32b_2746}, and
\underline{\textbf{(22)}} XiYanSQL-QwenCoder-32B~\cite{XiYanSQL}, the top four
open-source models on the BIRD leaderboard~\cite{bird-leaderboard}\footnote{Based
on the BIRD single-model leaderboard as of 2026-08-24.}.
We further evaluate the two best-performing open-source agents on the Spider
2.0-lite leaderboard~\cite{spider-2-leaderboard}\footnote{Based on the Spider
2.0-lite leaderboard as of 2026-08-24.}: \underline{\textbf{(23)}}
ktx~\cite{kaelio2026ktx} paired with Codex~\cite{openai2026codexcli} using
GPT-5.5, and \underline{\textbf{(24)}} Databao Agent~\cite{databao_agent} using
GPT-5.2; each using the strongest backbone LLM available at its leaderboard submission.
\end{itemize}

\input{figures/posttrain-settings}


\minihead{Execution environment: access to data management tools.}
We evaluate all models both with and without specialized tools.
\begin{itemize}[leftmargin=*]
    \item \underline{No tools}: this is the default setup commonly used by AI agents, where models have access to a single generic coding tool (Section~\ref{subsec:tools}), which allows them to interact with the environment by executing generated code over the input tables.
    \item \underline{With tools}: in addition to the generic coding tool, models are also provided with \sys's specialized data management tools (\code{search}, \code{transform}, and \code{join}) illustrated in Section~\ref{subsec:tools}.
\end{itemize}

\minihead{Execution environment: coding languages.}
\noindent For the coding tool, we test all models with two common language setups: SQL and Python, to ensure the generalizability of our results.

\begin{itemize}[leftmargin=*]
    \item \underline{SQL}:
all data files are loaded into a SQLite database.
Models are provided with the list of table names and can interact directly with the database through SQL queries.
The outputs of every \texttt{SELECT} statement are returned to the model as 
execution feedback for the next iteration.
\item \underline{Python}:
all data files are loaded as CSV files in a temporary working directory.
Models are provided with file paths and can write Python code to manipulate data.
All outputs generated during execution are returned to the model as feedback.
\end{itemize}

\minihead{Evaluation metrics.}
Our metric of model quality is accuracy. 
Given a benchmark query $q$, we consider a model's output table $O$ correct if it matches the ground-truth table $R$ in table shape (enforcing the correct aggregation granularity) and cell-by-cell values (allowing for row and column permutation).
Our evaluation follows the same general convention used in prior NL2SQL-style evaluations \cite{spider2}, in aspects such as numeric rounding and formatting comparisons. 
\iftoggle{full}
{
    Specifically: \underline{Numeric rounding and formatting}.
    For numeric cells, we first normalize common display formatting, including currency symbols, percentage suffixes, and datetime parsing, and then compare parsed numeric values. Following Spider2,\footnote{\href{https://github.com/xlang-ai/Spider2/blob/main/spider2-dbt/evaluation_suite/eval_utils.py\#L81}{Spider2 evaluation script line 81}} we apply numeric tolerance during result comparison where
    two numeric values $a$ and $b$ are considered matched if
    $
    \texttt{np.isclose}(a,b,\texttt{rtol}=10^{-2},\texttt{atol}=0.5)
    $.
    \underline{Null and NaN handling}.
    Following Spider2,\footnote{\href{https://github.com/xlang-ai/Spider2/blob/main/spider2-dbt/evaluation_suite/eval_utils.py\#L121}{Spider2 evaluation script line 121}} we canonicalize missing values before comparison. Values such as \texttt{None}, \texttt{NaN}, empty strings, \texttt{"null"}, and \texttt{"n/a"} are treated as missing.
    \underline{Aggregation granularity and table shape}.
    We require the predicted and ground-truth dataframes to have the same shape before cell-level matching, except for scalar ground truth, where a matching scalar value in the prediction is accepted.
}
{
    Details of the evaluation protocol are documented in our technical report~\cite{full}.
}

We report all results averaged over 10 repeated runs to account for model randomness.
To assess statistical significance of accuracy gains, we apply a paired $t$-test over the 100 \bench tasks: for each pair of conditions being compared (e.g., with-tools vs.\ no-tools for the same model), we compute a per-task difference in success rate over the 10 runs, then test whether the mean is significantly greater than zero. We report 95\% confidence intervals calculated by paired bootstrap (10{,}000 resamples over tasks). 

We also report model costs, based on the number of input/output tokens, and publicly available pricing information.
\footnote{All pricing information is based on data retrieved in 2026-08.
For large proprietary models, we use their official API pricing~\cite{openai_api_pricing}.
For Llama4 Maverick, we use the Azure AI pricing~\cite{cloudprice_llama4_maverick_pricing}.
For Llama3-8B, we use the Lambda pricing~\cite{cloudprice_llama3p1_8b_instruct_pricing}.
For Qwen3-8B and its variants, we use the LlamaGate pricing~\cite{cloudprice_qwen3_8b_pricing}.
For Infly-RL-SQL-32B, we use the Fireworks AI pricing~\cite{deepseek_r1_distill_qwen_32b_fireworks}.
For XiYanSQL-QwenCoder-32B, we use the OVHcloud pricing~\cite{qwen25_coder_32b_instruct_ovhcloud}.
For Mistral-Large-3, we use the Fireworks AI pricing~\cite{cloudprice_mistral_large_3}.
For DeepSeek-V4-Pro, we use the Fireworks AI pricing~\cite{cloudprice_deepseek_v4_pro}.
For Kimi-K2.6, we use the Cloudflare Workers AI pricing~\cite{cloudflare_kimi_k2p6}.
For Kwai-AutoSQL-32B and Kwai-AutoSQL-14B, we use the Fireworks AI pricing of their corresponding Qwen3-32B and Qwen3-14B backbones~\cite{cloudprice_qwen3_32b,cloudprice_qwen3_14b}.
    We report the average cost required to execute each query once.}


\subsection{Main Results: Quality and Costs}
\label{subsec:overall_res}

\input{figures/tool_v_no_tool_v2}
\input{figures/posttrain_update_v3}
\minihead{\sys's data management tools greatly improve quality.}
Table~\ref{tab:main_res_tool} shows that across all models and under both SQL and Python settings, \sys's data management tools (described in Section \ref{subsec:tools}) provide substantial quality improvements, on average increasing accuracy by 14 percentage points for SQL and 11 percentage points for Python. We report that these improvements are statistically significant in 19 out of 20 cases.\footnote{
As a robustness check for within-project dependence, we repeated all
significance analyses with projects as the unit of analysis. We
averaged the per-task differences within each of the 82 unique projects
and recomputed the $t$-tests and bootstrap CIs over the resulting
project-level differences. Every gain in Tables \ref{tab:main_res_tool} and  \ref{tab:main_res_post_train} 
significant at the task level
remained significant at the project level.
To ensure our significant gains are not false positives arising from the number of comparisons, we controlled the false discovery rate at $q{=}0.05$ across all 36 pairwise comparisons in the two tables using the Benjamini--Hochberg procedure~\cite{benjamini1995controlling}; the set of significant gains was unchanged.
}

While the benefit is substantial across the board, it is most pronounced for GPT-OSS-120B, whose accuracy rises by over 8$\times$ under both SQL and Python, as it is powerful in coding but struggles to process complex data in \bench without the help of tools.
\iftoggle{full}
{
    Specifically, without \sys's tools,
    GPT-OSS-120B barely interacts with the data, terminating after a median of
    3 model calls under SQL and 2 under Python without inspecting value encodings, join keys, or the
    distributions of the columns it aggregates. \sys's tools give it a concrete way
    to interrogate the data before committing to an answer, and the model makes substantial use of them: the
    median call count rises to 9 under SQL and 5 under Python, and each of those
    calls replaces a guess about the data with an observation of it.

}


These consistent gains demonstrate that \sys's reference tool design allows LLMs to offload difficult data processing tasks to specialized data management methods, delivering strong quality improvements for both highly capable frontier models and smaller open-source models in complex BI workflows.

\minihead{\sys post-training brings substantial additional gains.}
Table~\ref{tab:main_res_post_train} shows that post-training (Section~\ref{subsec:training}) can substantially enhance Qwen3-8B's capability for this in-domain task of end-to-end BI, with all overall gains statistically significant at $p{<}0.001$.\footnote{Both task and project levels.}

Specifically, SFT models improve over the base model by 17.4 percentage points without tools 
and 22 percentage points with tools using SQL. 
RL further improves the SFT models in all tool and language settings: when combined with \sys's tools, Qwen3-8B post-trained with RL improves by 29.8 percentage points in SQL. 
Notably, this performance is \textit{comparable to large models, even surpassing GPT-4o and 4 out of the 5 large open-source models} in their no-tools settings.
We will show that these gains also transfer to out-of-domain tasks in
Section~\ref{subsec:data_ablation}.

\minihead{Post-trained \sys models greatly reduce cost.}
Table~\ref{tab:main_res_post_train} shows that our post-trained \sys models achieve performance comparable to much larger models while dramatically reducing serving cost. For example, the post-trained Qwen3-8B-RL model costs only \$0.19 to run the entire \bench, whereas large models such as o4-mini and GPT-4o can cost over \$10, making our post-trained \sys models $54\times$ more cost-effective while 
delivering similar performance.

In terms of the total cost of ownership, our total training cost of small 8B models is less than \$200\footnote{We post-trained all Qwen3-8B variants on $2\times$ 80GB H100 GPUs, with SFT taking approximately 9 hours and RL 22 hours. Using public pricing info of \$2.89/GPU-hour~\cite{gpu-price}, this total training cost is less than \$200.}. Since Power BI and Tableau together serve over 100K organizations worldwide~\cite{bi-usage-1, bi-usage-2}, even if amortized over only one query per organization, the training cost would amount to less than \$0.002 per query, negligible compared with the roughly \$0.1 per-query inference
cost of frontier models.





\minihead{Existing SOTA NL2SQL methods are insufficient for \bench.} 
Table \ref{tab:main_res_post_train} shows that even top NL2SQL models, such as Kwai-AutoSQL-14B, Kwai-AutoSQL-32B, Infly-RL-SQL-32B and XiYanSQL-QwenCoder-32B, which are the top 4 open-source models on the BIRD leaderboard with over 70\% accuracy, produce dramatically lower quality scores of less than 10\% on \bench on average.
For coding agents, ktx and Databao Agent, the two best-performing open-source
agents on Spider 2.0-lite with over 70\% accuracy, also drop to around 25\% on
\bench.

We believe these results highlight aspects of the complexity and end-to-end nature of \bench that are not traditionally captured by NL2SQL, making \bench a challenging new benchmark and a valuable complement to existing NL2SQL research.

\subsection{Sensitivity Analysis}

\minihead{Quality improves steadily with more data management tools.}
In Figure~\ref{fig:ablation}, we examine the result quality when the number of enabled tools increases from 1 to 3.
Because the number of possible tool combinations is large, we simplify the presentation to highlight overall trends. Using GPT-4o as a representative model, for the 1-tool setting we report the average accuracy across the three single-tool configurations (\code{transform}, \code{search}, and \code{join}); for the 2-tool setting, we report the average across all $\binom{3}{2}$ tool combinations.


Figure~\ref{fig:ablation} shows that result quality improves monotonically as more data management tools are enabled for both SQL and Python. Moreover, the marginal gains increase with each additional tool. In Python, gains rise from 4.1 to 4.6 to 5.2 percentage points as one, two, and three tools are enabled; in SQL, they similarly increase from 1.8 to 3.7 to 5.1, which indicates that the tools are ``better together'' rather than independent. For example, \code{search} narrows the set of potentially relevant tables, making \code{join} more effective, and \code{transform} standardizes relational tables, which in turn makes both \code{search} and \code{join} easier. We observe that each tool makes a distinct contribution: enabling any additional tool improves quality, while removing any tool degrades it.
%
\newlength{\bucketrow}
\setlength{\bucketrow}{0.66\textwidth}     
\newlength{\ablwidth}
\setlength{\ablwidth}{0.46\bucketrow}    

\begin{figure*}[t]
%
\graphicspath{{figures/}}
\centering
\begin{minipage}[b]{\ablwidth}
    \centering
    \includegraphics[width=\linewidth]{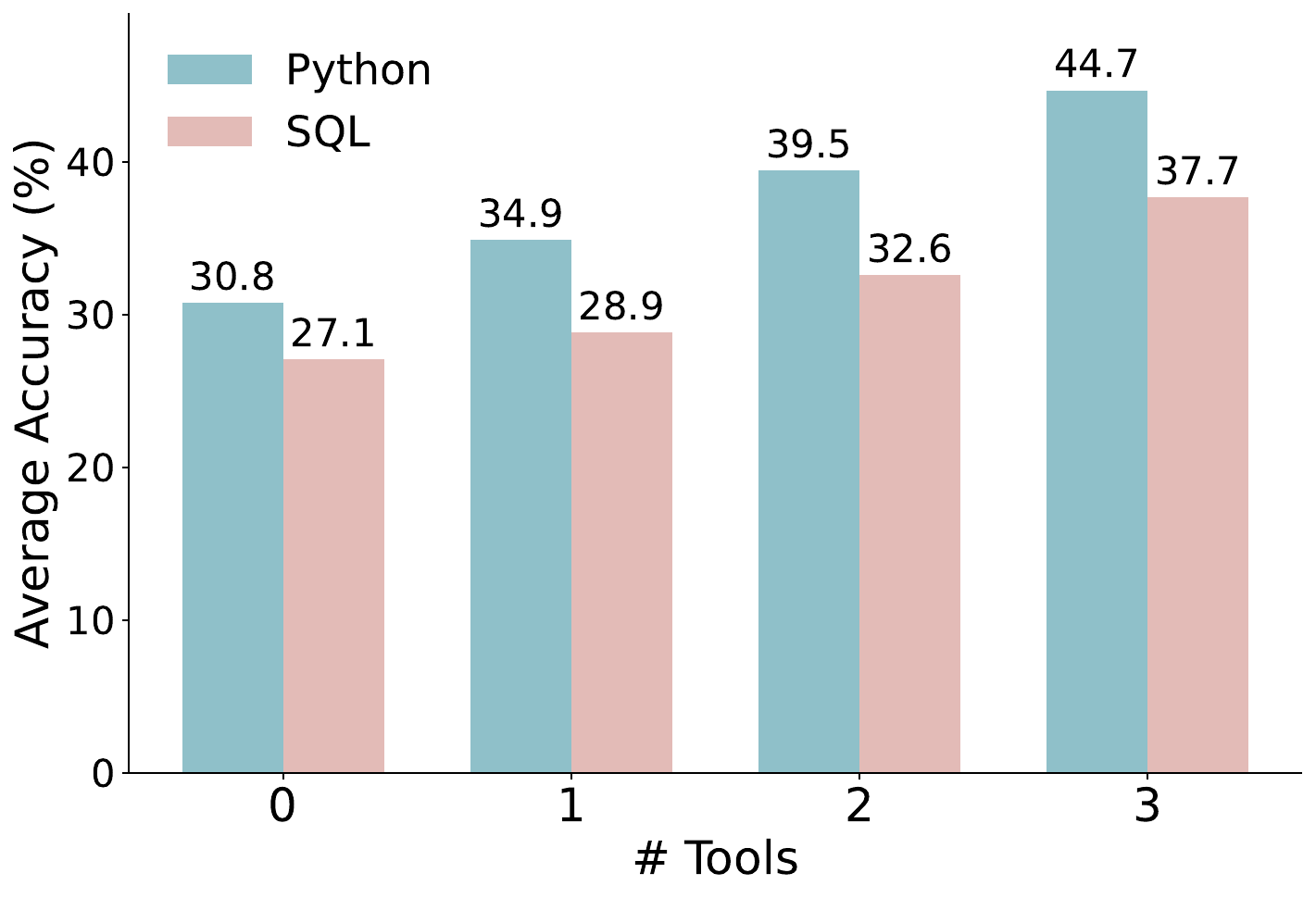}
    \caption{Quality of GPT-4o using different numbers of tools.}
    \label{fig:ablation}
\end{minipage}
\hfill
\begin{minipage}[b]{\bucketrow}
    \centering
    \begin{subfigure}[b]{0.311\linewidth}
        \centering
        \includegraphics[width=\linewidth]{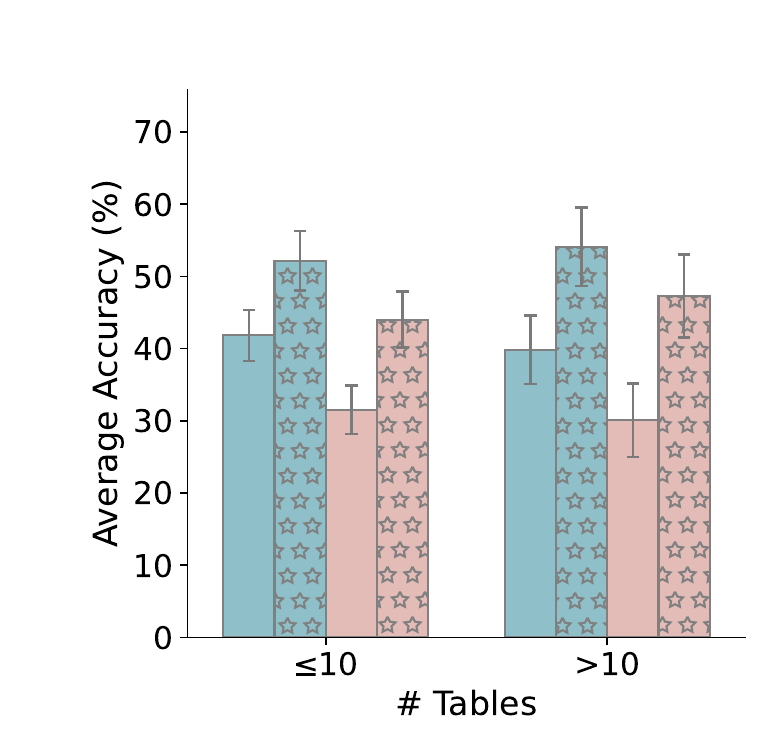}
    \end{subfigure}%
    \begin{subfigure}[b]{0.284\linewidth}
        \centering
        \includegraphics[width=\linewidth]{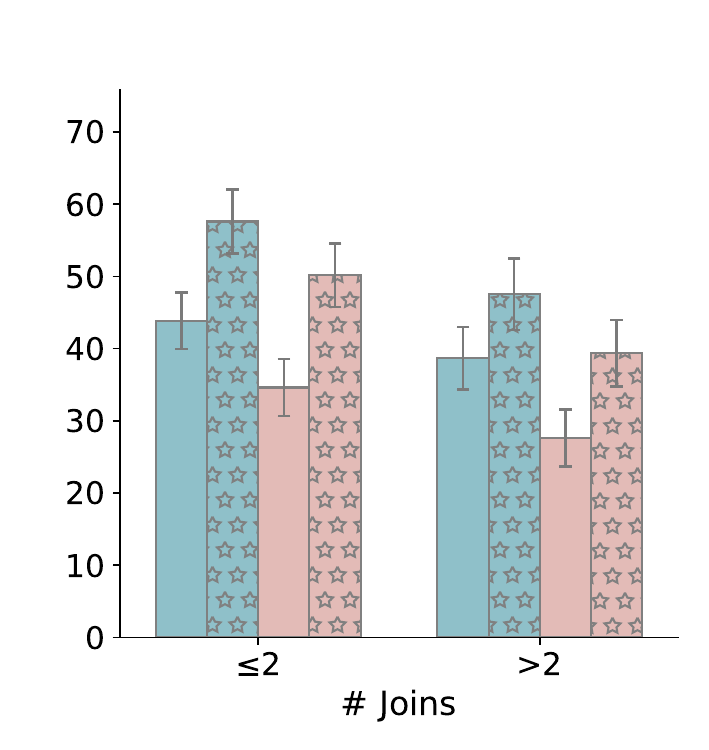}
    \end{subfigure}%
    \begin{subfigure}[b]{0.405\linewidth}
        \centering
        \includegraphics[width=\linewidth]{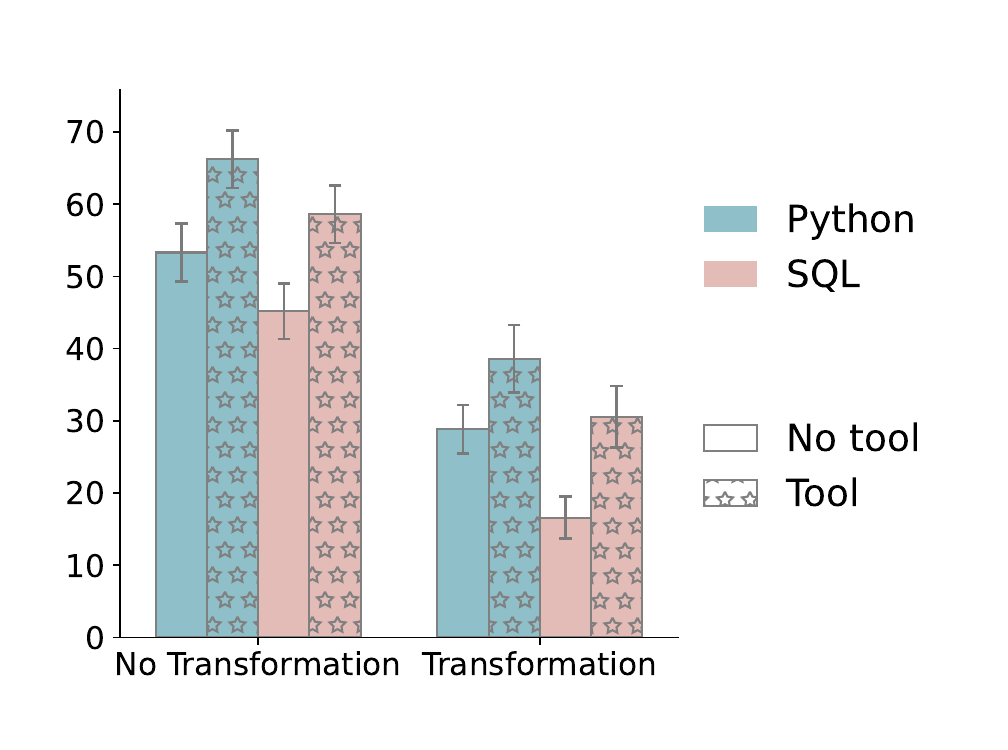}
    \end{subfigure}
    \caption{
    Model quality improvements when using tools, with tasks grouped by
    complexity: (left) number of tables, (middle) number of joins, and
    (right) transformations. Error bars are one standard error of the mean
    across \bench tasks.
    }
    \label{fig:complexity_breakdown_tool}
\end{minipage}
\end{figure*}





\begin{figure*}[t]
\centering

\begin{subfigure}[t]{0.315\textwidth}
    \centering
    \includegraphics[width=\linewidth]{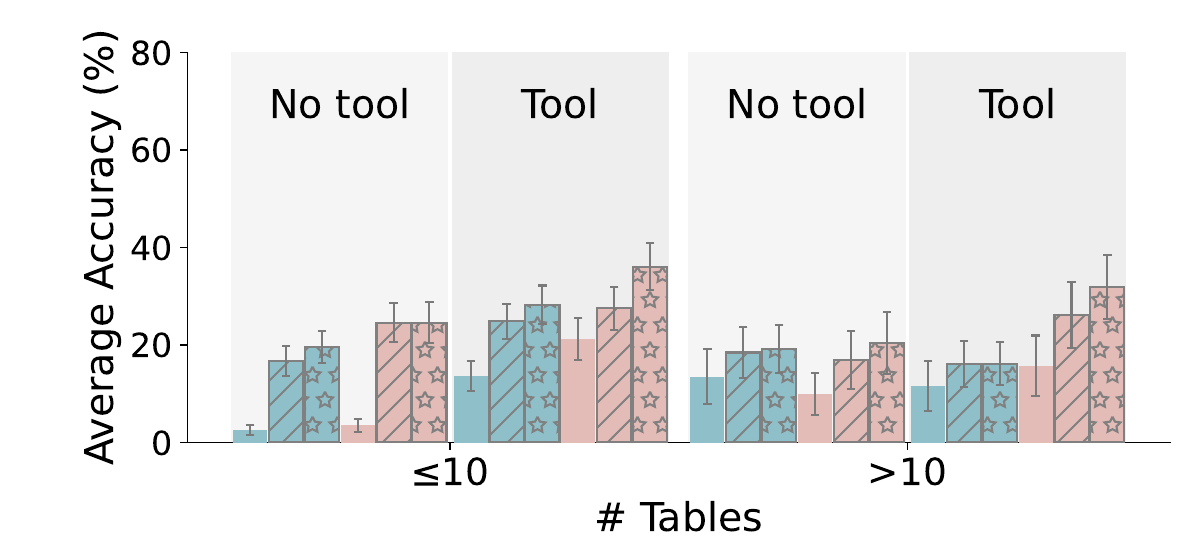}
\end{subfigure}
\hfill
\begin{subfigure}[t]{0.295\textwidth}
    \centering
    \includegraphics[width=\linewidth]{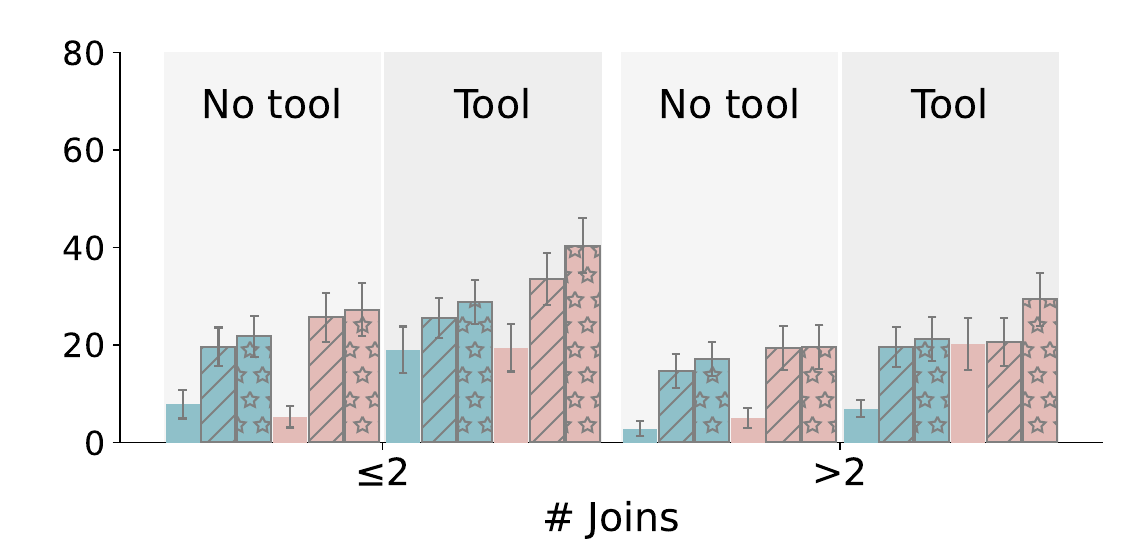}
\end{subfigure}
\hfill
\begin{subfigure}[t]{0.38\textwidth}
    \centering
    \includegraphics[width=\linewidth]{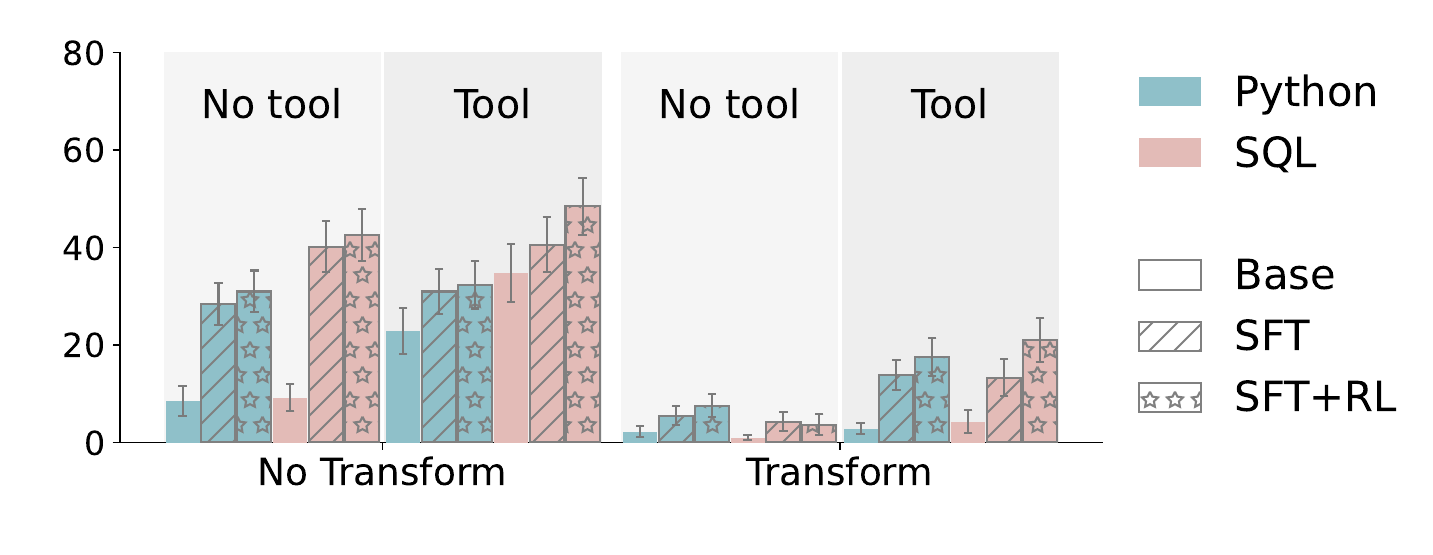}
\end{subfigure}
\caption{
Post-training quality gains (with and without tools), by task complexity.
We group tasks by (left) number of tables, (middle) number of joins, and (right) transformations.
Error bars are one standard error of the mean across \bench tasks.
}
\label{fig:complexity_breakdown_posttrain}
\end{figure*}

\minihead{\sys's tools provide robust gains across task complexities.}
Figure~\ref{fig:complexity_breakdown_tool} shows that data management tools produce consistent gains over different levels of task complexity, as measured by the number of tables, joins, and transformations. We note that the improvements are more pronounced for more complex tasks. For example, Figure~\ref{fig:complexity_breakdown_tool}(left) shows that on tasks requiring over 10 tables, using tools produces significantly higher gains (57.1\% for SQL and 35.8\% for Python) than on simpler tasks requiring fewer than 10 tables (39.5\% for SQL and 24.7\% for Python). Similarly, in Figure~\ref{fig:complexity_breakdown_tool}(right), for tasks requiring reshaping transformations, using tools produces relative quality gains of 84.2\% for SQL and 33.9\% for Python, which are also significantly higher than when no such transformations are required (29.7\% for SQL and 24.2\% for Python).

\minihead{\sys's post-training generates larger gains on more complex tasks.}
Figure~\ref{fig:complexity_breakdown_posttrain} shows that post-trained models achieve significant quality gains across the board when tasks are grouped by the number of tables, joins, and transformations, with SFT models always significantly outperforming base models and RL models further improving on SFT models. We observe that the quality gains also increase with task complexity. For example, Figure \ref{fig:complexity_breakdown_posttrain}(middle) shows that in the Python setting without data management tools, SFT improves accuracy by 1.5$\times$ and RL by 1.8$\times$ on tasks with at most 2 joins; these improvements increase to 4.1$\times$ and 5$\times$, respectively, on tasks with more than 2 joins. 


\minihead{\sys's reward design provides clear and informative learning signals for RL post-training.}
To validate our reward design for \sys's post-training (Section~\ref{subsec:training}), we conduct an ablation analysis in which we \underline{(1)} remove the $-0.5$ partial credit assigned to trajectories that return
a non-empty but incorrect result table;
\underline{(2)} remove the $-0.1n$ syntax-error penalty; and
\underline{(3)} remove both terms, which reduces the reward to a binary $\pm1$
signal. We evaluate these reward variants on the (Python, tool) setting:
starting from the SFT initialization (Qwen3-8B-SFT-Tool in Table~\ref{tab:main_res_post_train}) which achieves 22.6\% accuracy, our full
reward design (Qwen3-8B-RL-Tool) improves accuracy to 25.1\%.
In contrast, variants \underline{(1)}, \underline{(2)}, and \underline{(3)} achieve only 20.7\%, 19.1\%, and
20.6\%, respectively. All three ablations underperform even the SFT initialization, demonstrating that the explicit syntax feedback and dense partial credit in \sys's reward design are crucial for effective RL post-training.

\iftoggle{full}
{
    \subsection{Error Analysis}
    \label{subsec:error}

    \begin{figure*}[t]
        \centering
        \begin{subfigure}[t]{0.48\textwidth}
            \centering
            \includegraphics[width=\linewidth]{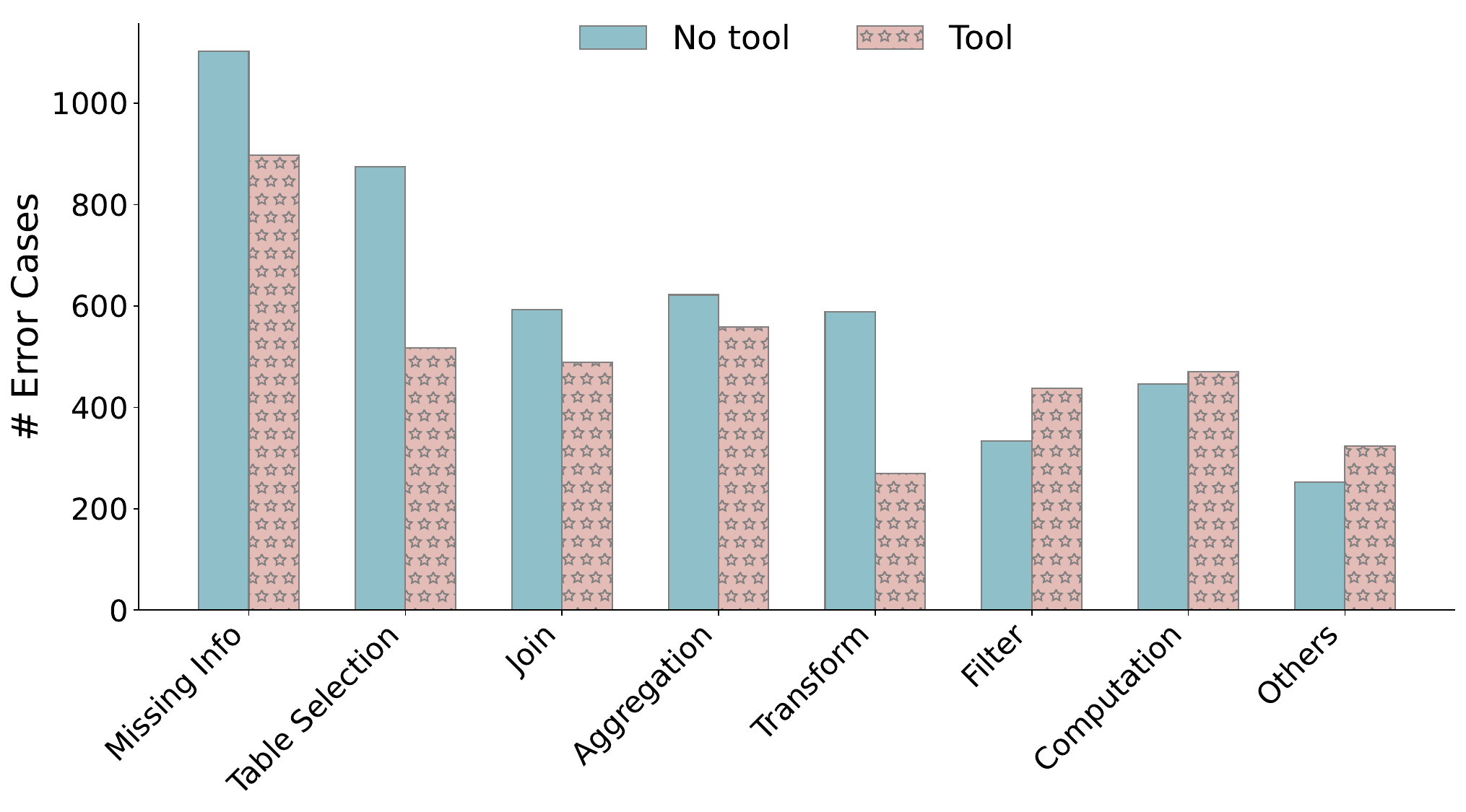}
            \caption{}
            \label{fig:errors}
        \end{subfigure}
        \hfill
        \begin{subfigure}[t]{0.48\textwidth}
            \centering
            \includegraphics[width=\linewidth]{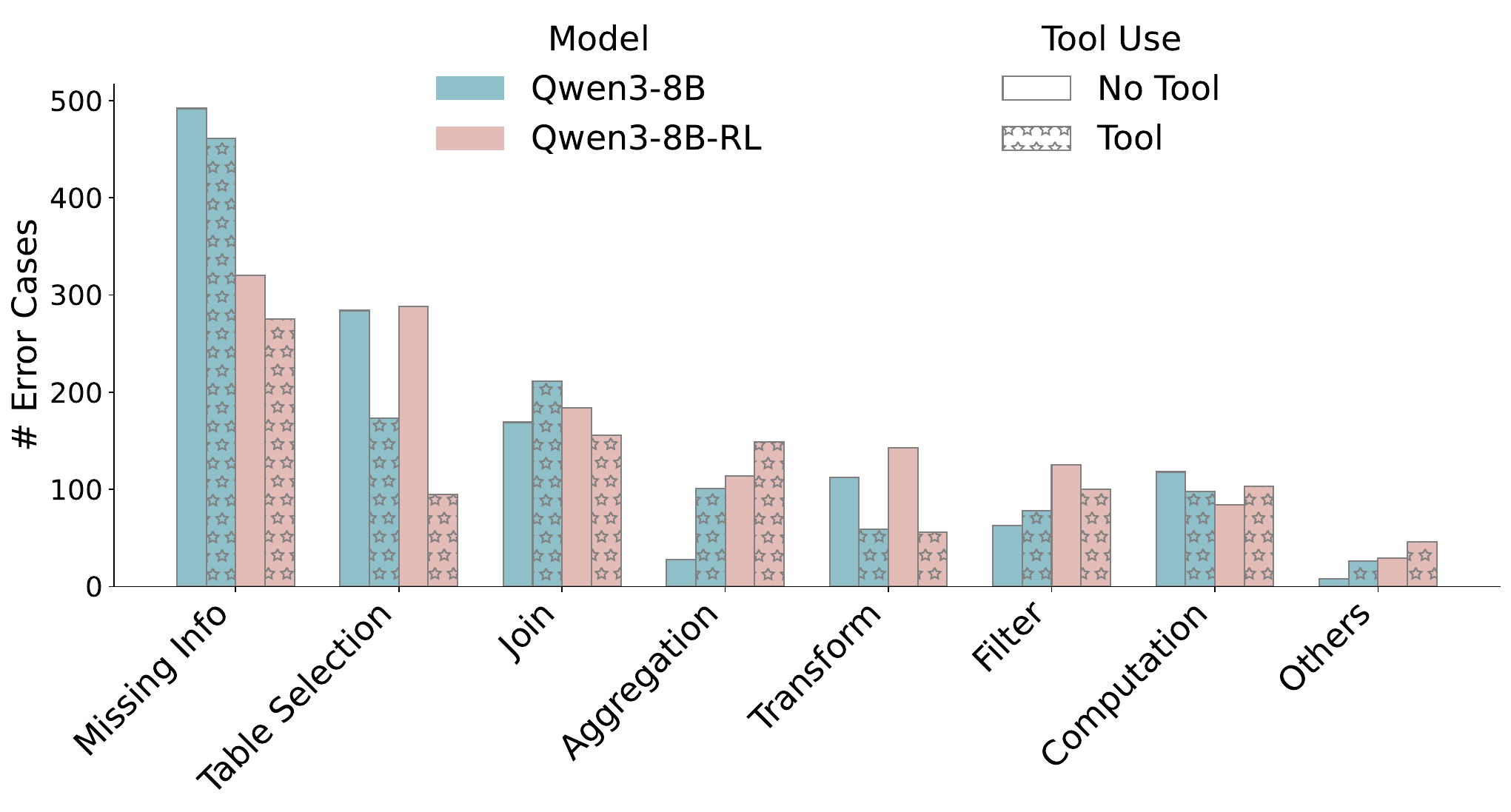}
            \caption{}
            \label{fig:errors_qwen}
        \end{subfigure}
        \caption{Error distributions for (a) tool vs. no-tool settings averaged over all large models and (b) base vs. post-trained Qwen3-8B.}
        \label{fig:error_distributions}
    \end{figure*}

    Figure~\ref{fig:error_distributions} shows a classification of error types based on model traces. Specifically, for each incorrect model trace, we compare it with a corresponding correct trace and ask GPT-5.2 to classify the primary root error into one of the following categories:
    \begin{itemize}[leftmargin=*]
      \item \underline{\textit{Transform Error}}: incorrect transformation of a table
            (e.g.\ pivot, unpivot, transpose).
      \item \underline{\textit{Join Error}}: incorrect combination of data across tables
            or sources: wrong join keys, missing joins, or misaligned entities.
      \item \underline{\textit{Table Selection Error}}: failure to select the tables
            needed to answer the query, either by choosing irrelevant tables or
            by omitting necessary ones.
      \item \underline{\textit{Computation Error}}: incorrect calculation or reasoning
            over data, e.g., arithmetic slips, logical errors,
            or misapplied formulas.
      \item \underline{\textit{Missing Information Error}}: failure to include required
            data, yielding incomplete output that omits key \emph{values, rows,
            or attributes}.
      \item \underline{\textit{Unnecessary Information Error}}: inclusion of extraneous
            data the query did not ask for.
      \item \underline{\textit{Filter Error}}: incorrect filtering conditions: a wrong
            subset, a missing condition, a superfluous constraint, or a wrong
            time range.
      \item \underline{\textit{Aggregation Error}}: incorrect computation over grouped
            data: wrong group-by keys, wrong aggregation function (e.g.\ sum
            v.s. average), or improper normalisation (e.g.\ wrong
            denominator).
      \item \underline{\textit{Deduplication Error}}: improper handling of duplicate
            records, either failing to remove duplicates or falsely removing necessary
            repeated entries.
      \item \underline{\textit{Temporal Error}}: incorrect handling of time: mixing
            periods, using outdated snapshots, or misaligning time ranges or
            granularities.
      \item \underline{\textit{Missing-Value Handling Error}}: incorrect treatment of
            nulls: improper removal, incorrect imputation, or treating a
            missing value as valid (e.g.\ as zero).
    \end{itemize}

    The categories are not mutually exclusive in principle -- a wrong join can
    also produce missing rows -- so the classifier is asked for the
    \emph{primary root} error, the earliest step whose correction would fix the
    trace.

    We report the error type distributions of all large models in Table \ref{tab:main_res_tool} in Figure \ref{fig:errors}. We can see that tools
    effectively reduce the most common error types. In particular, transform errors are reduced the most (by 54.2\%), followed by table selection errors (by 40.9\%). This demonstrates the effectiveness of our tool design.

    In Figure \ref{fig:errors_qwen} we compare the error type distributions of vanilla vs. post-trained Qwen3-8B,  under both tool and no-tool settings. Consistent with our findings on larger models, tools reduce the most common error types for both Qwen3-8B and Qwen3-8B-RL. In particular, for Qwen3-8B-RL, table selection errors decrease by 67\% and transformation errors decrease by 61\%. For Qwen3-8B, table selection errors decrease by 39\% and transformation errors decrease by 47\%. We also observe that post-training reduces several of the most common error types. In particular, table selection errors decrease by 16\% and join errors decrease by 11\%. Finally, tools and post-training complement each other and together achieve substantial reductions in error rates. From Qwen3-8B without tools to Qwen3-8B-RL with tools, table selection errors decrease by 67\% and transformation errors decrease by 50\%.

    In Figure \ref{fig:error_distributions}, we see that missing information remains the most common error type. These errors arise when a model fails to include required data needed to fully answer the query, leading to incomplete outputs with missing key values, rows, or attributes, which is an important direction for future work. 
}
{
     \minihead{Additional results.} We report a detailed error analysis (where we find join/transform/search and actual coding all account for significant fractions of the observed errors),  as well as  additional sensitivity analysis, in a full version of our paper~\cite{full}.
}

\subsection{Generalizability Test}
\label{subsec:data_ablation}
So far we have focused on \bench. To study whether our methods have generalizable benefits beyond BI (on out-of-domain tasks), we perform additional evaluations on Spider 2.0~\cite{spider2}, a challenging NL2SQL benchmark that spans diverse domains and topics.
\iftoggle{full}
{
    \footnote{We use the Spider 2.0-lite variant, which can be executed locally via SQLite.
    We select all 122 queries that are marked as local (i.e., without requiring Snowflake or BigQuery in the cloud) and do not require external knowledge. 
    }
}
We emphasize that \emph{Spider 2.0 is not only a completely new and unseen dataset, but also a very different task (NL2SQL),} which therefore serves as a strong generalizability test for \sys.


Table~\ref{tab:spider_res} compares the quality of the base Qwen3-8B with that of the Qwen3-8B-SFT and Qwen3-8B-RL models post-trained in \sys (Section~\ref{subsec:training}) on Spider 2.0. Our findings of \sys on \bench generalize reliably to Spider 2.0 for both tool use and model post-training. Tools consistently improve over the corresponding no-tool settings, SFT models outperform the base model, and RL models further improve over SFT models, even though both SFT and RL post-training were performed on entirely different BI tasks.

We note that by combining tools and SFT+RL post-training, the Qwen3-8B-RL model achieves a $20.5\times$ quality improvement over the base model's accuracy, 
underscoring the generalizability of \sys to new and unseen tasks beyond \bench.

\input{figures/spider_res}

\iftoggle{full}
{
    \subsection{Cost and latency benefits}
    \label{subsec:rounds}




    \begin{figure}[t]
        \centering

        \begin{subfigure}[t]{0.32\columnwidth}
            \centering
            \includegraphics[width=\linewidth]{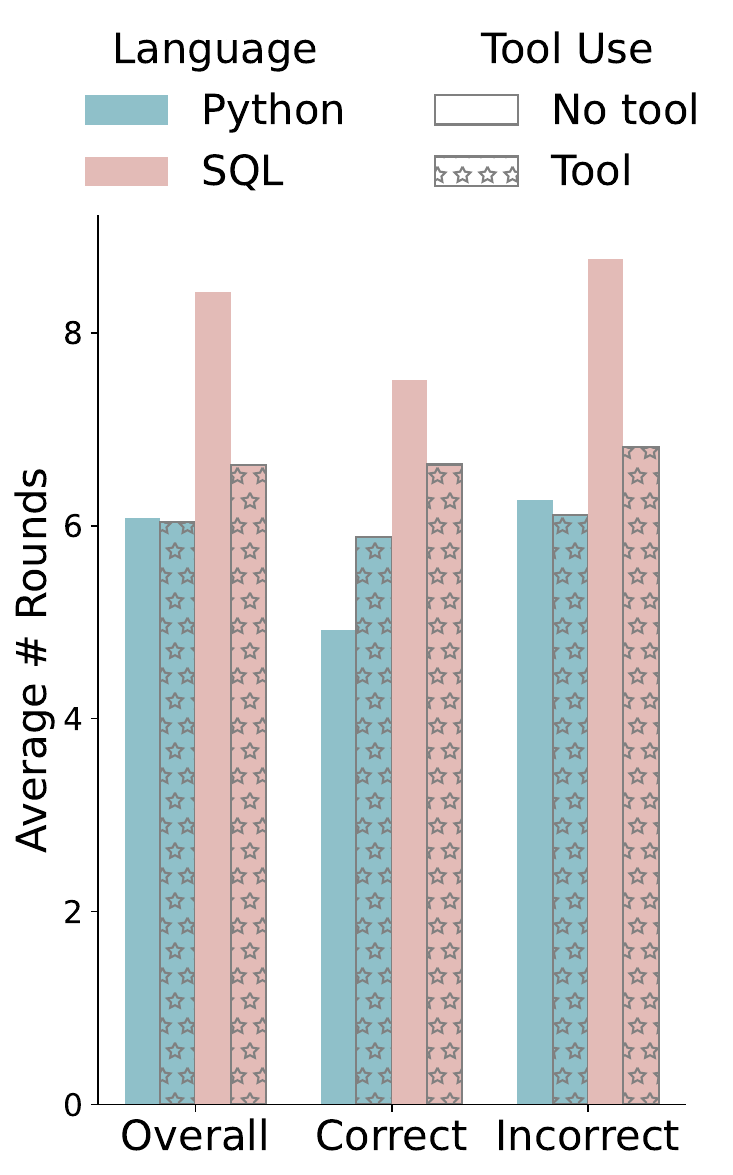}
            \caption{}
            \label{fig:round_tool}
        \end{subfigure}
                \hfill
        \begin{subfigure}[t]{0.32\columnwidth}
            \centering
            \includegraphics[width=\linewidth]{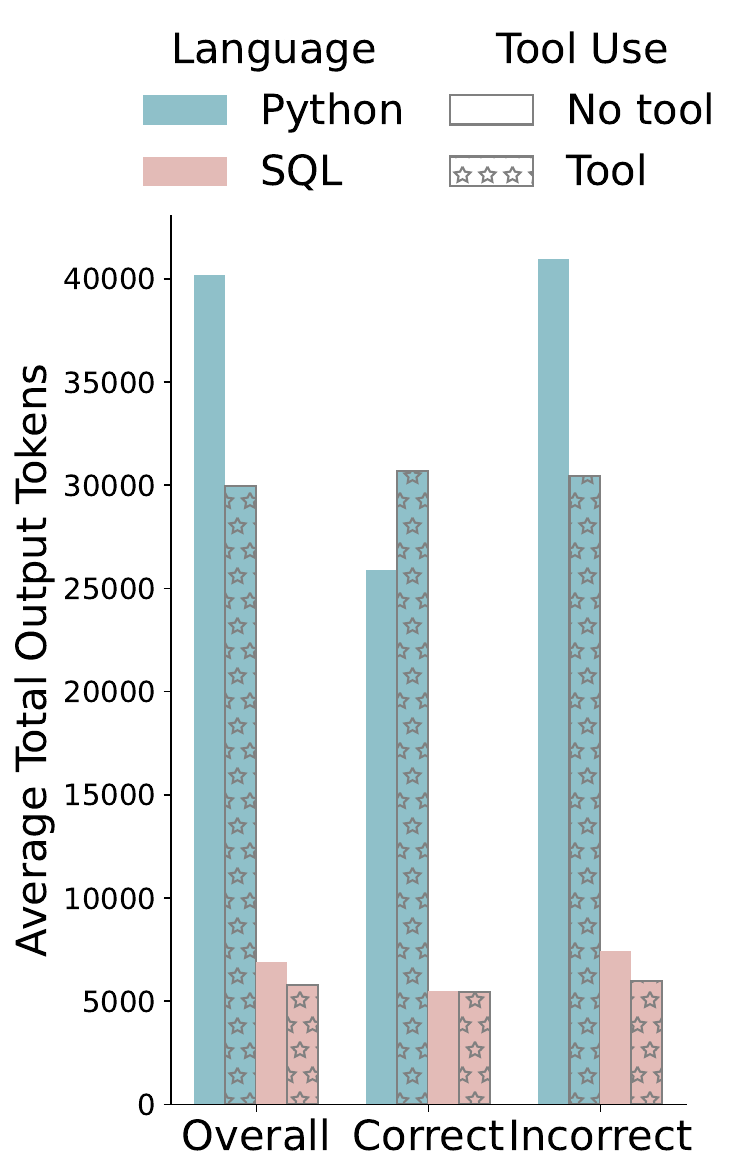}
            \caption{}
            \label{fig:token_total}
        \end{subfigure}
        \hfill
        \begin{subfigure}[t]{0.32\columnwidth}
            \centering
            \includegraphics[width=\linewidth]{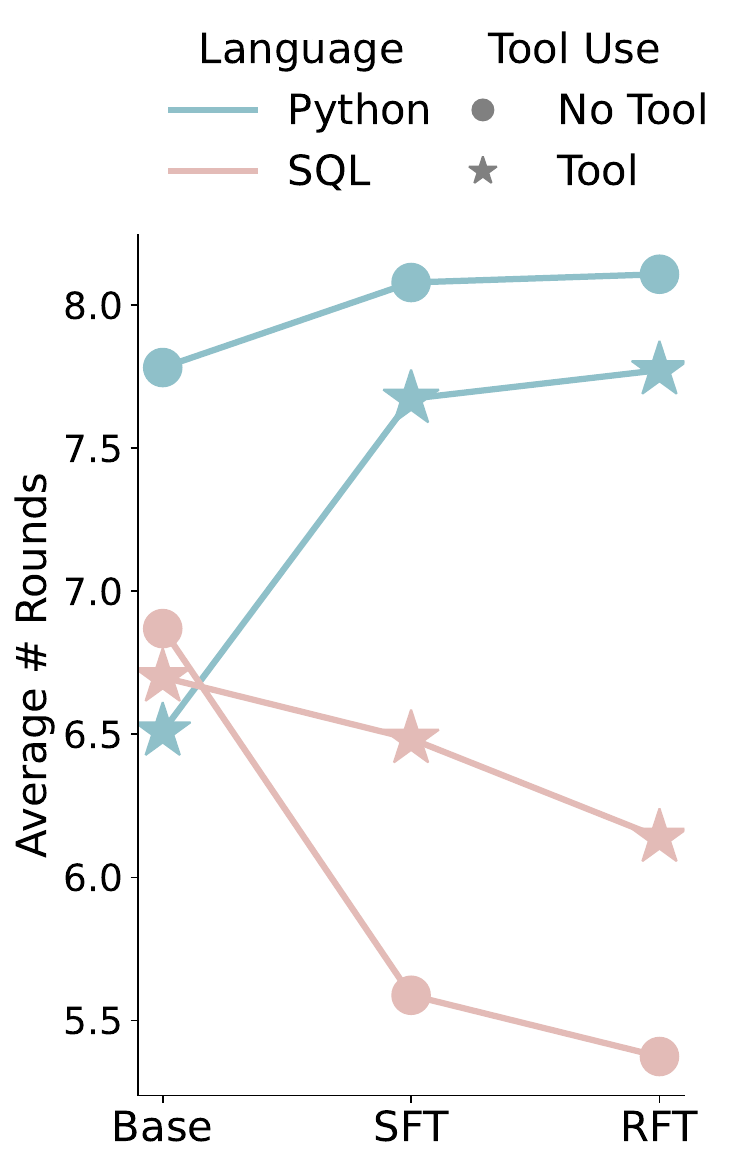}
            \caption{}
            \label{fig:round_train}
        \end{subfigure}

        \caption{
    Number of interaction rounds (a) and total output tokens (b) required to complete a task on \bench under different settings, across all models (overall, correct, and incorrect executions); and interaction rounds across post-training stages of Qwen3-8B (c).
        }
        \label{fig:round_and_tokens}
    \end{figure}

    We additionally study the cost and latency benefits of \sys through tool use and post-training. Figure \ref{fig:round_and_tokens} shows the number of interaction rounds for \sys under different settings. Our tool design and post-training pipeline optimize these interactions. Specifically, we have the following findings.

    \minihead{\tools reduce interaction rounds}
    As we show in Figure~\ref{fig:round_tool}, incorporating tools reduces the average number of interaction rounds overall, decreasing from 7.2 to 6.3, even though introducing 3 additional tools could theoretically add 3 extra interaction steps. 
    This indicates that the tools replace longer multi-step reasoning or coding processes with single, structured operations.
    By directly executing complex BI operations, the tools eliminate redundant exploration and reduce unnecessary debugging, leading to more efficient and stable interaction trajectories.

    The reduction is more pronounced for SQL than for Python: for SQL, tool usage decreases the average number of interaction rounds from 8.4 to 6.6, whereas for Python the decrease is more modest, from 6.1 to 6.0.
    This is because, as we mentioned in Section \ref{subsec:overall_res}, certain complex transformation operations, such as transpose, require multiple iterative refinement steps without tools in SQL; in contrast, Python offers native data manipulation libraries with expressive built-in operations for such transformations, leading to comparatively smaller gains from additional tool usage.

    For correct executions, tools slightly increase the average number of interaction rounds from 6.2 without tools to 6.3 with tools, whereas for incorrect executions, tools reduce the average rounds substantially from 7.5 without tools to 6.5 with tools. This indicates that tools help prevent prolonged failure modes by enabling faster recovery or earlier termination when the model is off track. They also introduce debugging and verification steps that can correct intermediate mistakes and eventually reach the correct solution.


    \minihead{Incorrect executions take more rounds.}
    As we show in Figure~\ref{fig:round_tool}, incorrect executions require more interaction rounds than correct ones. On average, correct executions take 6.4 rounds, while incorrect executions take 7.1 rounds. This trend is consistent across both Python and SQL. The reason is that once a trajectory deviates from the correct reasoning path, the model often requires additional debugging, retrying, or compensatory steps before termination. In contrast, successful executions typically follow a more coherent reasoning trajectory with fewer corrective detours, leading to shorter interaction sequences.

    \minihead{\tools decrease user observed latency.}
    As we can see from Figure \ref{fig:token_total}, the total number of output tokens decreases by 25\% for Python and 16\% for SQL when tools are applied. Combined with our findings from Figure~\ref{fig:round_tool} that the number of interaction rounds also decreases, we conclude that \tools can meaningfully reduce user-observed latency and model response time.

    \minihead{SFT increases exploration, while RL improves efficiency}
    As we show in Figure~\ref{fig:round_train}, the average number of interaction rounds initially increases after SFT and then decreases following RL. This is because during SFT, training on traces generated by large models encourages more active tool use and explicit intermediate reasoning, resulting in longer interaction trajectories. RL further optimizes the policy using rollout-based feedback, enabling the model to refine its action strategy and interact with the data more efficiently, thereby reducing unnecessary steps.
} 
{
    \minihead{Additional results.} We report additional findings, such as how tool use and post-training in \sys reduce models' output tokens (therefore costs), and the number of interaction rounds (which reduces user-observed latency), in a full version of our paper~\cite{full}.
} 

%% file: figures/posttrain-settings.tex





\begin{table*}[t]
\vspace{-10mm}
\centering
\caption{A total of 8 post-trained model variants of Qwen3-8B, using (SFT, SFT+RL) $\times$ (No tools, Tools) $\times$  (SQL, Python).
}
\resizebox{0.8\textwidth}{!}{%
\begin{tabular}{lcccc}
\toprule
 & \multicolumn{2}{c}{SQL} & \multicolumn{2}{c}{Python} \\
\cmidrule(lr){2-3} \cmidrule(lr){4-5}
Qwen3-8B Variants & no-tools & with-tools & no-tools & with-tools \\
\midrule

SFT 
& \underline{\textbf{(11)}} Qwen3-8B-SFT (SQL)
& \underline{\textbf{(12)}} Qwen3-8B-SFT-Tool (SQL)
& \underline{\textbf{(13)}} Qwen3-8B-SFT (Python)
& \underline{\textbf{(14)}} Qwen3-8B-SFT-Tool (Python) \\

SFT + RL 
& \underline{\textbf{(15)}} Qwen3-8B-RL (SQL)
& \underline{\textbf{(16)}} Qwen3-8B-RL-Tool (SQL)
& \underline{\textbf{(17)}} Qwen3-8B-RL (Python)
& \underline{\textbf{(18)}} Qwen3-8B-RL-Tool (Python) \\

\bottomrule
\end{tabular}
}
\label{tab:exp_settings}
\end{table*}

%% file: figures/tool_v_no_tool_v2.tex
\begin{table*}[t]
\caption{
\sys's data management tools can greatly improve quality for end-to-end BI: Accuracy (\%) and cost per query (\$0.01) comparison across different models on \bench. We annotate quality gains with \allimp{+x}. Gains marked \sigstar{$^{***}$} ($p{<}0.001$), \sigstar{$^{**}$} ($p{<}0.01$), and \sigstar{$^{*}$} ($p{<}0.05$) are statistically significant (paired $t$-test, $n{=}100$); 95\% CIs
\ci{[l,r]} are obtained from paired bootstrap.
}
\centering
\resizebox{0.8\textwidth}{!}{
\begin{tabular}{p{2.4cm}cc cc cc cc}
\toprule
& \multicolumn{4}{c}{SQL} & \multicolumn{4}{c}{Python} \\
\cmidrule(lr){2-5} \cmidrule(lr){6-9}
Models
& \multicolumn{2}{c}{no-tools} & \multicolumn{2}{c}{with-tools}
& \multicolumn{2}{c}{no-tools} & \multicolumn{2}{c}{with-tools} \\
\cmidrule(lr){2-3} \cmidrule(lr){4-5} \cmidrule(lr){6-7} \cmidrule(lr){8-9}
& Accuracy$\uparrow$ & Cost$\downarrow$
& Accuracy$\uparrow$ & Cost$\downarrow$
& Accuracy$\uparrow$ & Cost$\downarrow$
& Accuracy$\uparrow$ & Cost$\downarrow$ \\
\midrule

o4-mini
& 48.2 & 10.3 & 61.9~\allimp{+13.7}\sigstar{$^{***}$}~\ci{[+7.7,+19.7]} & 7.64
& 57.4 & 3.95 & 66.3~\allimp{+8.9}\sigstar{$^{*}$}~\ci{[+1.8,+15.8]} & 5.61 \\
GPT-5.5
& 46.7 & 7.17 & 60.2~\allimp{+13.5}\sigstar{$^{***}$}~\ci{[+8.1,+19.4]} & 14.91
& 59.5 & 8.66 & 65.4~\allimp{+5.9}\sigstar{$^{*}$}~\ci{[+1.0,+11.0]} & 16.97 \\
GPT-5.2
& 46.1 & 3.03 & 55.2~\allimp{+9.1}\sigstar{$^{***}$}~\ci{[+4.4,+14.2]} & 5.77
& 48.6 & 3.08 & 60.2~\allimp{+11.6}\sigstar{$^{***}$}~\ci{[+6.1,+17.1]} & 5.58 \\
Mistral-Large-3
& 36.2 & 4.81 & 44.8~\allimp{+8.6}\sigstar{$^{**}$}~\ci{[+2.7,+14.7]} & 6.06
& 35.0 & 3.23 & 51.1~\allimp{+16.1}\sigstar{$^{***}$}~\ci{[+10.0,+22.5]} & 6.20 \\
Llama-4-Maverick
& 27.5 & 9.91 & 31.2~\allimp{+3.7}~\ci{[-4.7,+11.9]} & 6.85
& 28.6 & 9.24 & 38.8~\allimp{+10.2}\sigstar{$^{***}$}~\ci{[+4.5,+16.2]} & 7.79 \\
GPT-4o
& 27.1 & 8.97 & 37.7~\allimp{+10.6}\sigstar{$^{***}$}~\ci{[+5.1,+16.1]} & 8.75
& 30.8 & 7.07 & 44.7~\allimp{+13.9}\sigstar{$^{***}$}~\ci{[+8.4,+19.9]} & 11.0 \\
DeepSeek-V4-Pro
& 23.5 & 2.99 & 30.8~\allimp{+7.3}\sigstar{$^{***}$}~\ci{[+3.4,+11.3]} & 1.71
& 53.2 & 3.57 & 57.6~\allimp{+4.4}\sigstar{$^{*}$}~\ci{[+0.8,+8.1]} & 3.37 \\
Kimi-K2.6
& 20.0 & 18.0 & 36.7~\allimp{+16.7}\sigstar{$^{***}$}~\ci{[+10.7,+22.9]} & 23.6
& 55.6 & 8.87 & 62.7~\allimp{+7.1}\sigstar{$^{**}$}~\ci{[+3.0,+11.4]} & 10.5 \\
GPT-OSS-120B
& 5.3 & 0.33 & 45.3~\allimp{+40.0}\sigstar{$^{***}$}~\ci{[+33.3,+46.8]} & 2.29
& 3.1 & 0.20 & 27.3~\allimp{+24.2}\sigstar{$^{***}$}~\ci{[+19.5,+29.2]} & 0.90 \\
Qwen3-8B
& 5.2 & 0.17 & 19.8~\allimp{+14.6}\sigstar{$^{***}$}~\ci{[+8.2,+21.3]} & 0.32
& 5.4 & 1.21 & 13.1~\allimp{+7.7}\sigstar{$^{*}$}~\ci{[+1.9,+13.9]} & 1.37 \\
\bottomrule
\end{tabular}
}
\label{tab:main_res_tool}
\end{table*}

%% file: figures/posttrain_update_v3.tex
\begin{table}[t]
\caption{
\sys's post-training framework significantly improves quality: Accuracy (\%) and cost per query (\$0.01) comparisons across different post-trained Qwen3-8B variants and SOTA NL2SQL methods and coding agents on \bench.
We annotate overall gains \allimp{+x}, and when applicable, break them down into tool gain (\toolimp{+x}) and post-training gain (\trainimp{+x}).
All gains marked \sigstar{$^{***}$} ($p{<}0.001$), \sigstar{$^{**}$} ($p{<}0.01$), and \sigstar{$^{*}$} ($p{<}0.05$) are statistically significant (paired $t$-test, $n{=}100$); 95\% CIs
\ci{[l,r]} are obtained from paired bootstrap.
}
\centering
\resizebox{\columnwidth}{!}{
\begin{tabular}{p{3.6cm}cc cc}
\toprule
& \multicolumn{2}{c}{SQL} & \multicolumn{2}{c}{Python} \\
\cmidrule(lr){2-3} \cmidrule(lr){4-5}
Models and Agents
& Accuracy$\uparrow$ & Cost$\downarrow$
& Accuracy$\uparrow$ & Cost$\downarrow$ \\
\midrule

\multicolumn{5}{l}{\emph{\textbf{\sys} }} \\

Qwen3-8B-SFT
& 22.6~\allimp{+17.4}\sigstar{$^{***}$}
& \multirow[c]{2}{*}{0.17}
& 17.2~\allimp{+11.8}\sigstar{$^{***}$}
& \multirow[c]{2}{*}{1.89} \\[-1pt]

\hspace{1.2em}\subrow{}
& \multicolumn{1}{c}{
    \subrow{\ci{[+11.9,+23.3]}}
  }
&
& \multicolumn{1}{c}{
    \subrow{\ci{[+7.3,+16.8]}}
  }
& \\[3pt]

Qwen3-8B-SFT-Tool
& 27.2~\allimp{+22.0}\sigstar{$^{***}$}
& \multirow[c]{3}{*}{0.31}
& 22.6~\allimp{+17.2}\sigstar{$^{***}$}
& \multirow[c]{3}{*}{1.61} \\[-2pt]

\hspace{1.2em}\subrow{}
& \multicolumn{1}{c}{
    \subrow{\ci{[+15.6,+29.1]}}
  }
&
& \multicolumn{1}{c}{
    \subrow{\ci{[+11.4,+23.3]}}
  }
& \\[-2pt]

\hspace{1.2em}\subrow{}
& \multicolumn{1}{c}{
    \subrow{%
      \begin{tabular}{@{}r@{\hspace{4pt}}l@{}}
        \toolimp{+4.6} & \trainimp{+7.4}\sigstar{$^{***}$} \\[-4pt]
        \cisub{[-1.1,+10.2]} & \cisub{[+3.6,+11.6]}
      \end{tabular}%
    }
  }
&
& \multicolumn{1}{c}{
    \subrow{%
      \begin{tabular}{@{}r@{\hspace{4pt}}l@{}}
        \toolimp{+5.4}\sigstar{$^{*}$} & \trainimp{+9.5}\sigstar{$^{***}$} \\[-4pt]
        \cisub{[+0.4,+10.6]} & \cisub{[+4.8,+14.5]}
      \end{tabular}%
    }
  }
& \\[4pt]

Qwen3-8B-RL
& 23.5~\allimp{+18.3}\sigstar{$^{***}$}
& \multirow[c]{2}{*}{0.19}
& 19.5~\allimp{+14.1}\sigstar{$^{***}$}
& \multirow[c]{2}{*}{1.97} \\[-1pt]

\hspace{1.2em}\subrow{}
& \multicolumn{1}{c}{
    \subrow{\ci{[+12.4,+24.6]}}
  }
&
& \multicolumn{1}{c}{
    \subrow{\ci{[+9.0,+19.5]}}
  }
& \\[3pt]

Qwen3-8B-RL-Tool
& \textbf{\underline{35.0}}~
  \allimp{+29.8}\sigstar{$^{***}$}
& \multirow[c]{3}{*}{0.31}
& \textbf{\underline{25.1}}~
  \allimp{+19.7}\sigstar{$^{***}$}
& \multirow[c]{3}{*}{1.70} \\[-2pt]

\hspace{1.2em}\subrow{}
& \multicolumn{1}{c}{
    \subrow{\ci{[+22.7,+37.1]}}
  }
&
& \multicolumn{1}{c}{
    \subrow{\ci{[+13.4,+26.3]}}
  }
& \\[-2pt]

\hspace{1.2em}\subrow{}
& \multicolumn{1}{c}{
    \subrow{%
      \begin{tabular}{@{}r@{\hspace{4pt}}l@{}}
        \toolimp{+11.5}\sigstar{$^{**}$} & \trainimp{+15.2}\sigstar{$^{***}$} \\[-4pt]
        \cisub{[+5.1,+18.1]} & \cisub{[+10.1,+20.8]}
      \end{tabular}%
    }
  }
&
& \multicolumn{1}{c}{
    \subrow{%
      \begin{tabular}{@{}r@{\hspace{4pt}}l@{}}
        \toolimp{+5.6} & \trainimp{+12.0}\sigstar{$^{***}$} \\[-4pt]
        \cisub{[-0.0,+11.4]} & \cisub{[+7.2,+17.0]}
      \end{tabular}%
    }
  }
& \\[4pt]

\midrule

{\textbf{\emph{SOTA NL2SQL methods}}} \\

ktx w/ Codex (GPT-5.5)
& 26.3 & 22.15
& - & - \\

Databao Agent (GPT-5.2)
& 23.8 & 5.54
& - & - \\

Kwai-AutoSQL-32B
& 17.3 & 3.52
& 9.2 & 3.86 \\

Kwai-AutoSQL-14B
& 7.8 & 0.74
& 4.0 & 0.97 \\

XiYanSQL-QwenCoder-32B
& 7.7 & 2.32
& 13.2 & 6.58 \\

Infly-RL-SQL-32B
& 6.0 & 2.23
& 5.5 & 3.23 \\





\bottomrule
\end{tabular}

}
\label{tab:main_res_post_train}
\end{table}

%% file: figures/spider_res.tex
\begin{table}[t]
\caption{
\sys's data management tools and post-training improve performance on Spider 2.0, an unseen NL2SQL benchmark (best results are  in bold and underlined). 
}
 \vspace{-1mm}
\centering
\resizebox{0.85\columnwidth}{!}{
\begin{tabular}{p{2.3cm}cccc}
\toprule
& \multicolumn{2}{c}{SQL} & \multicolumn{2}{c}{Python} \\
\cmidrule(lr){2-3} \cmidrule(lr){4-5}
Models & no-tools & with-tools & no-tools & with-tools \\
\midrule

\hspace{0.3em} Qwen3-8B
& 0.8 & 4.9 & 0 & 1.6 \\

\hspace{0.3em} Qwen3-8B-SFT
& 10.7
& 13.1
& 2.5
& 4.9 \\

\hspace{0.3em} Qwen3-8B-RL
& 13.1
& \textbf{\underline{16.4}}
& 3.3
& \textbf{\underline{8.2}}\\
\bottomrule
\end{tabular}
}
\label{tab:spider_res}
\end{table}

%% file: tex/7-Conclusions.tex
\section{Conclusions and Future Work}
In this work, we take a first step towards answering ad-hoc BI questions with LLMs. We build \bench, the first benchmark to systematically evaluate LLMs using BI workflows, and propose \sys, an agentic framework for end-to-end BI, where data-management tools and model post-training lead to substantial improvements. We believe LLM-driven BI is a promising direction, and hope our work can serve as a springboard for future research.